\documentclass[a4paper,10pt]{article}
\usepackage[margin=1in]{geometry} 
\usepackage[utf8]{inputenc} 
\usepackage[T1]{fontenc}
\usepackage{graphicx} 
\usepackage{amsmath, amssymb} 
\usepackage{float} 
\usepackage{hyperref}
\usepackage{authblk} 
\usepackage{setspace}
\usepackage{caption}
\usepackage{xcolor}
\title{\textbf{\textcolor{black}{\textsc{Dynami-CAL GraphNet}}: A Physics-Informed Graph Neural Network Conserving Linear and Angular Momentum for Dynamical Systems}}

\author[1]{Vinay Sharma}
\author[1]{Olga Fink}

\affil[1]{Intelligent Maintenance and Operations Systems, EPFL, Lausanne, Switzerland}

\date{}

\begin{document}

\maketitle

\section*{Abstract}
Accurate, interpretable, and real-time modeling of multi-body dynamical systems is essential for predicting behaviors and inferring physical properties in natural and engineered environments. Traditional physics-based models face scalability challenges and are computationally demanding, while data-driven approaches like Graph Neural Networks (GNNs) often lack physical consistency, interpretability, and generalization. In this paper, we propose \textcolor{black}{
\textsc{Dynami-CAL GraphNet}}, a Physics-Informed Graph Neural Network that integrates the learning capabilities of GNNs with physics-based inductive biases to address these limitations. \textcolor{black}{\textsc{Dynami-CAL GraphNet}} enforces pairwise conservation of linear and angular momentum for interacting nodes using edge-local reference frames that are equivariant to rotational symmetries, invariant to translations, and equivariant to node permutations. This design ensures physically consistent predictions of node dynamics while offering interpretable, edge-wise linear and angular impulses resulting from pairwise interactions. Evaluated on a 3D granular system with inelastic collisions, \textcolor{black}{\textsc{Dynami-CAL GraphNet}} demonstrates stable error accumulation over extended rollouts, effective extrapolations to unseen configurations, and robust handling of heterogeneous interactions and external forces. \textcolor{black}{\textsc{Dynami-CAL GraphNet}} offers significant advantages in fields requiring accurate, interpretable, and real-time modeling of complex multi-body dynamical systems, such as robotics, aerospace engineering, and materials science. By providing physically consistent and scalable predictions that adhere to fundamental conservation laws, it enables the inference of forces and moments while efficiently handling heterogeneous interactions and external forces. This makes it invaluable for designing control systems, optimizing mechanical processes, and analyzing dynamic behaviors in both natural and engineered systems. 

\section{Introduction}
\label{sec:intro}
Dynamical systems are fundamental to both natural and engineered environments, encompassing phenomena such as granular flows, molecular dynamics, and planetary motions in nature, as well as engineered components like bearings, gearboxes, and suspension systems. Accurately modeling these systems is essential for predicting behaviors and informing design, optimization, and operational management decisions. In operational settings, reliable models are especially valuable for learning from observational data, tracking state evolution in \textcolor{black}{real time}, and inferring key physical quantities that influence system dynamics \cite{jimenez2020towards}. However, developing physics-grounded models with explicit parametric differential equations requires a deep understanding of underlying \textcolor{black}{mechanics—posing challenges for complex systems with unknown interaction laws or unmeasurable parameters.} 
\textcolor{black}{Moreover, for many systems,} the computational \textcolor{black}{cost of numerical simulation} hinders real-time deployment \cite{an2015practical}, motivating the need for alternative approaches that learn interpretable dynamical models directly from observed data \textcolor{black}{and enable fast inference in operational} settings. 



\textcolor{black}{To achieve these advantages, data-driven models that learn dynamics from trajectory data have become popular, including surrogate models for predictive maintenance~\cite{zhang2019data}, control~\cite{brunton2022data}, and efficient multi-body simulations~\cite{choi2021data}. However, these models often lack physical consistency and 
generalize poorly beyond training conditions, learning spurious patterns tied to the training distribution~\cite{karniadakis2021physics}. Additionally, they suffer from error accumulation during rollout, leading to poor long-term predictions. Moreover, they typically require large training datasets—feasible for small systems with simple dynamics but prohibitive in complex multiphysics simulations (e.g., Direct Numerical Simulation of fluid flow) or real-world scenarios with limited, costly measurements, such as full-body motion capture or internal loads in rotating machinery.}

\textcolor{black}{Physics-Informed Neural Networks (PINNs)~\cite{raissi2019physics} address data scarcity and promote physical consistency by enforcing a learning bias during training. This involves using governing equations as additional constraints alongside data. However, they are sensitive to hyperparameters, expensive to train~\cite{krishnapriyan2021characterizing}, and face challenges in complex systems like multi-body dynamics, where enforcing constraints at every step becomes difficult (e.g., modeling a 9-segment human walker requires 17 nonlinear constraints~\cite{hu2004human}). Moreover, governing equations often rely on simplified assumptions to model effects like nonlinear friction, resulting in suboptimal performance compared to data-driven alternatives \cite{peng2022mechanistic}. }

Graph Neural Networks (GNNs) offer a flexible alternative for learning the dynamics of physical systems by embedding inductive biases into their architecture. Spatial inductive bias--representing components as nodes and interactions as edges--enables GNNs to learn the dynamics of physical systems via message passing, as demonstrated by the Graph Neural Simulator (GNS)~\cite{sanchez2020learning}. \textcolor{black}{GNNs have since been applied across domains including molecular dynamics~\cite{atz2021geometric}, granular flows~\cite{choi2024graph}, and engineered  systems such as bearings~\cite{sharma2023graph}.} However, models relying solely on spatial inductive bias often struggle to maintain physical consistency, leading to error accumulation in long rollouts and poor generalization to unseen conditions~\cite{han2022learning}.

\textcolor{black}{Another important inductive bias in physical modeling is symmetry--specifically, equivariance to 3D translations and rotations--reflecting the principle that physical laws are independent of the observer’s coordinate frame \cite{schutt2018schnet,satorras2021n}. Equivariant-GNNs incorporate this inductive bias, enabling improved modeling of physical systems \cite{han2025survey}. Broadly, Equivariant-GNNs fall into two classes: (i) scalarization-vectorization approaches that operate directly in 3D space, and (ii) high-degree steerable models that lift features to higher-order representations using spherical harmonics.}

\textcolor{black}{In the scalarization–vectorization paradigm, directional messages are generated by first computing scalar edge embeddings from node and edge features (scalarization), and then using them to scale geometric vectors such as relative positions (vectorization). For example, E(n)-Equivariant Graph Neural Networks (EGNNs)~\cite{satorras2021n} follow this approach by modulating relative position vectors with learned weights. Building on this, Graph Mechanics Networks (GMN)~\cite{huang2022equivariant} extend single channel message passing by incorporating multiple geometric channels--e.g., relative position and velocity vectors—each scaled independently to capture richer interactions. Further, ClofNet~\cite{du2022se} introduces equivariant edge-local reference frames, using projected node features to produce richer scalar embeddings that generate learned coefficients to modulate local basis vectors—enabling directional encoding beyond what relative vectors offer. Equivariant Graph Hierarchical Networks (EGHN)~\cite{han2022equivariant} build on GMN by incorporating hierarchical message passing to capture multi-scale dynamics. Other methods in this paradigm include Radial Field Networks (RF)~\cite{kohler2019equivariant} and SchNet~\cite{schutt2018schnet}.}

\textcolor{black}{The second class of high-degree steerable models includes methods that encode steerable features using spherical harmonics, transform them under rotations via Wigner-D matrices, and fuse them through Clebsch–Gordan tensor products—achieving full SE(3) equivariance (i.e., equivariance to the Special Euclidean group in 3D, encompassing both rotations and translations) at a higher computational cost \cite{han2025survey}. Representative examples include Tensor Field Networks (TFNs) \cite{thomas2018tensor}, SE(3)-Transformers \cite{fuchs2020se}, Neural Equivariant Interatomic Potentials (NequIP) \cite{batzner20223}, and Steerable E(3) Equivariant Graph Neural Networks (SEGNNs) \cite{brandstetter2022geometric}.}

\textcolor{black}{Symmetry-based inductive biases—such as translation and rotation equivariance—have substantially advanced the capability of GNNs to model the dynamics of physical systems. However, these symmetries alone  do not always  guarantee  that the learned models will respect fundamental physical laws. Several approaches have sought to embed hard physical priors by enforcing energy conservation, such as in Hamiltonian and Lagrangian GNNs~\cite{DBLP:journals/corr/abs-1909-12790, bhattoo2022learning}. However, these methods tend to underperform in settings with dissipation or external forces unless such effects are explicitly modeled~\cite{gruver2022deconstructing,sosanya2022dissipative}. Recently, ~\cite{han2022learning} further showed that incorporating Hamiltonian structure into EGNN degraded performance on a dynamics prediction task including external force.}

\textcolor{black}{In contrast to energy conservation, Newton’s third law guarantees that internal pairwise interactions conserve linear and angular momentum, even in the presence of dissipation or external forcing. In this work, we propose a principled method to integrate these universal inductive biases -- conservation of \textbf{linear} and \textbf{angular momentum} -- directly into the proposed equivariant GNN architecture. By embedding these conservation laws as structural biases within the network, we ensure that the model’s predictions consistently respect these key physical principles.}

\textcolor{black}{While some existing models (e.g., RF~\cite{kohler2019equivariant}, EGNN~\cite{satorras2021n}, GMN~\cite{huang2022equivariant}, ClofNet~\cite{du2022se}) can conserve linear momentum under certain  architectural constraints, this property is often lost in practice. For instance, RF, EGNN, and GMN form  edge embeddings as \( m_{ij} = \phi(Z^TZ, h_i, h_j) \), where \( Z \) encodes  relative geometric features (such as \( \vec{x}_{ij} \) or \( \vec{v}_{ij} \)). Incorporating  node features \( h_i \), \( h_j \) for expressivity  often results in non-symmetric embeddings (\( m_{ij} \ne m_{ji} \)), which, when used to scale antisymmetric relative vectors (e.g., \( \psi(m_{ij}) \cdot \vec{x}_{ij} \)), lead to non-antisymmetric forces (\( \vec{f}_{ij} \ne -\vec{f}_{ji} \)) and violate the net force cancellation required for linear momentum conservation. ClofNet introduces more expressive, non-relative edge embeddings (due to features like \( \vec{x}_i \times \vec{x}_j \) in addition to relative \( \vec{x}_{ij} \)) by projecting geometric features onto an edge-local reference frame. However, it inherits the same node dependence (\( m_{ij} = \phi(Z^T Z, h_i, h_j) \)) and thus fails to ensure \( m_{ij} = m_{ji} \). Moreover, its orthonormal basis \( (\vec{a}, \vec{b}, \vec{c}) \), defined as \( \vec{a}_{ij} = \hat{x}_{ij} \), \( \vec{b}_{ij} = \hat{x}_i \times \hat{x}_j \), and \( \vec{c}_{ij} = \vec{a}_{ij} \times \vec{b}_{ij} \), is not fully antisymmetric under node interchange: \( \vec{a}_{ij} = -\vec{a}_{ji} \), \( \vec{b}_{ij} = -\vec{b}_{ji} \), but \( \vec{c}_{ij} = \vec{c}_{ji} \), which again breaks the antisymmetry needed for force conservation. Other approaches, such as Flux-GNN~\cite{horie2024graph} and Conservation-informed GNN~\cite{DBLP:journals/corr/abs-2412-20962}, preserve flux symmetry in scalar conservation laws using permutation-invariant constructions (e.g., DeepSets\cite{NIPS2017_f22e4747}), ensuring \( m_{ij} = m_{ji} \). However, both are designed for scalar partial differential equations (PDEs): FluxGNN relies on radial vectors (normals to cells) and cannot capture non-central forces—a drawback shared with EGNN—while CiGNN lacks rotational equivariance. Furthermore, these methods construct edge embeddings solely from relative features, which can limit their expressivity when modeling complex directional interactions.}

\textcolor{black}{Conserving angular momentum poses an even greater challenge. For two bodies with moments of inertia \( I_i \), masses \( m_i \), angular velocities \(\vec{\omega_i} \), and linear velocities \( \vec{v}_i \)—interacting via equal and opposite but non-central forces (i.e., not aligned with the vector connecting their centers), the total angular momentum about a reference point \( \vec{r}_0 \) is given by both spin \( \sum I_i \omega_i \) and orbital \( \sum (\vec{r}_i - \vec{r}_0) \times m_i \vec{v}_i \) contributions. Non-central forces alter the orbital component of angular momentum by changing linear momentum, necessitating compensatory rotational torques to preserve total angular momentum. These torques—arising from force–moment arm interactions or pure couples—are neither symmetric nor antisymmetric and are not explicitly modeled in prior GNN architectures. While forces govern changes in translational degrees of freedom, it is these torques that drive the evolution of the rotational state.}

\textcolor{black}{To address these limitations, we propose \textbf{\textsc{Dynami-CAL GraphNet}}—a \textsc{Dynami}cs-predictor \textsc{Graph} neural \textsc{Net}work that explicitly conserves angular and linear momentum by embedding these conservation laws directly into the model architecture. This approach  enables physically consistent predictions even under complex, non-central, and dissipative interactions, 
while remaining applicable across diverse systems. As a six-degree-of-freedom model, \textsc{Dynami-CAL GraphNet} predicts both internal forces and rotational torques through three key innovations:
\textbf{(1) Novel edge-local reference frames.} We introduce a novel edge-aligned orthonormal basis that is equivariant to 3D rotations (SO(3)), invariant to translations (T(3)), and antisymmetric under node exchange—ensuring equal and opposite internal forces in accordance with Newton’s third law. Node vector features (e.g., velocity, angular velocity) are projected onto this basis and combined with node and edge scalar features to form expressive invariant edge embeddings during scalarization.
\textbf{(2) Novel physically grounded vectorization.} Edge embeddings are decoded into three vector channels: (i) an antisymmetric internal force vector (enforcing linear momentum exchange), (ii) a pairwise angular interaction vector (governing total angular momentum exchange), and (iii) a predicted force application point. The spin torque---responsible for angular velocity updates---is  computed by isolating the orbital contribution  (via the cross product of force and lever arm) from the angular interaction vector, treating each edge as a self-contained dynamical system.
\textbf{(3) Spatiotemporal message passing.} Our message-passing scheme incorporates  sub-time stepping, enabling edge embeddings to accumulate information across both spatial neighbors and previous iterations. This design facilitates fine-grained dynamic modeling and robust generalization across diverse physical systems.}

\textcolor{black}{Overall, \textsc{Dynami-CAL GraphNet} advances the field by directly embedding the core conservation principles of classical mechanics into the model architecture, enabling accurate and generalizable predictions for the dynamics of complex, real-world systems.}

\begin{figure}[h]  
\centering \includegraphics[width=0.95\textwidth]{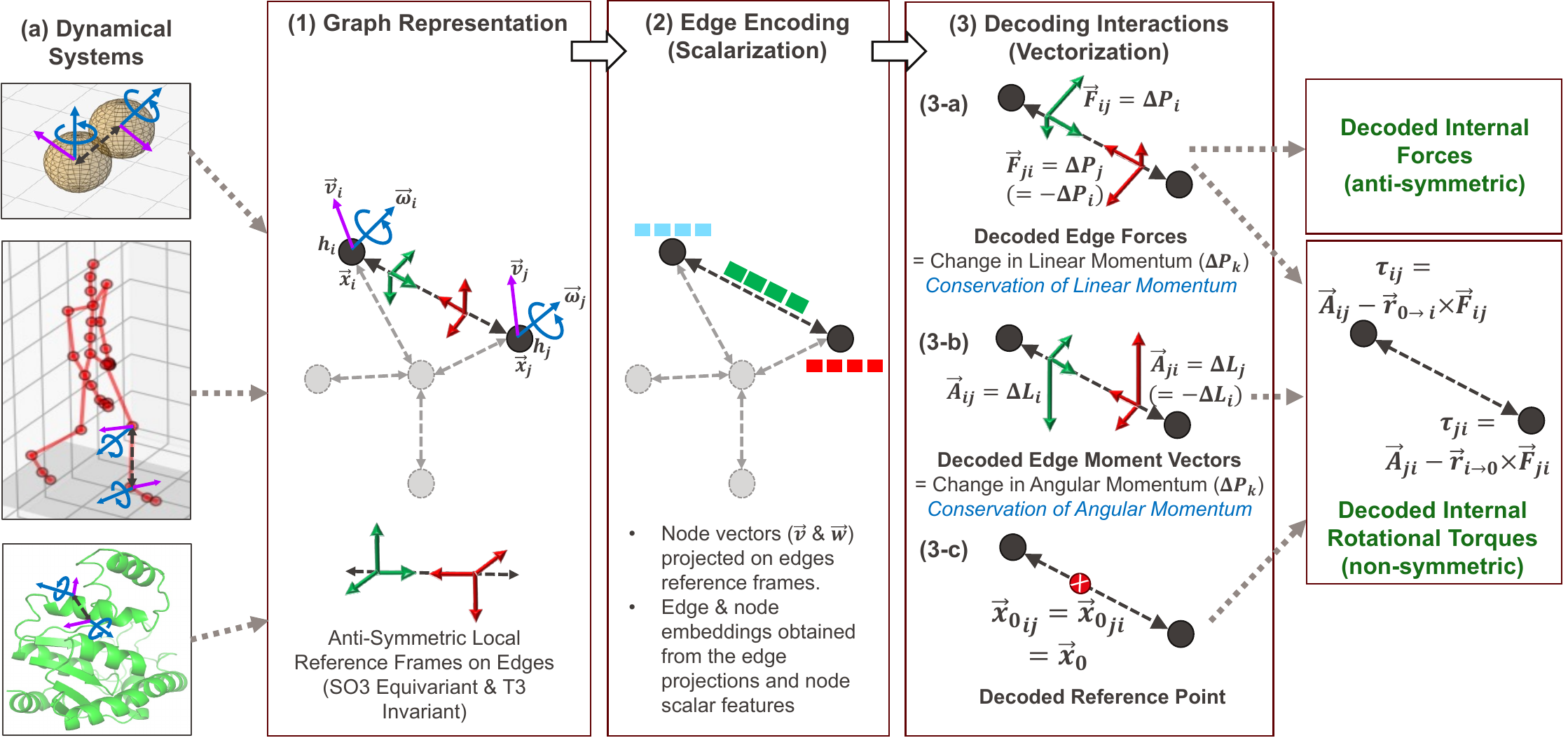}  
\caption{\textcolor{black}{
\textbf{Conservation of Linear and Angular Momentum in \textsc{Dynami-CAL GraphNet}.}  
\textsc{Dynami-CAL GraphNet} incorporates symmetry- and conservation-based inductive biases to model complex dynamical systems such as granular collisions, human motion, and molecular dynamics (a).  
\textbf{(1) Graph Representation:} Each edge is equipped with an SO(3)-equivariant, T(3)-invariant, and antisymmetric local reference frame, enforcing symmetry under node interchange.  
\textbf{(2) Scalarization: Edge Embedding} Node vector features (e.g., $\vec{v}$, $\vec{\omega}$) are projected into these frames and combined with scalar features to form invariant scalar edge embeddings.  
\textbf{(3) Vectorization: Decoding Interactions}  
(3-a) Embeddings are decoded into antisymmetric internal forces ($\vec{F}_{ij} = -\vec{F}_{ji}$), conserving linear momentum.  
(3-b) Decoded Antisymmetric angular momentum changes ($\vec{A}_{ij} = -\vec{A}_{ji}$) ensure total angular momentum conservation.  
(3-c) The predicted point of force application enables isolation of spin torque via subtraction of the orbital component, yielding non-symmetric torques that update angular velocity.  
}}
\label{fig:plcons}  
\end{figure}

\subsubsection*{\textsc{\textcolor{black}{Dynami-CAL GraphNet}}}
\textcolor{black}{\textsc{Dynami-CAL GraphNet} is a general framework for modeling six-degree-of-freedom (6-DoF) dynamics  in complex physical systems  using a structured scalarization–vectorization pipeline. The model is highly versatile, accommodating a wide range of systems—granular assemblies, biomolecules, and articulated human motion—by representing them as graphs. In this formulation,   nodes encode position, linear velocity \( \vec{v}_i \), and angular velocity \( \vec{\omega}_i \), while  bi-directional edges represent pairwise interactions between system components. The overall architecture and data flow are illustrated in Fig.~\ref{fig:plcons}, highlighting how this graph-based approach enables flexible and accurate modeling of complex dynamical behaviors across diverse domains.}

\textcolor{black}{Each edge is assigned  a local orthonormal reference frame that is equivariant to 3D rotations (SO(3)), invariant to translations (T(3)), and antisymmetric under node interchange (Fig.~\ref{fig:plcons}–1). In practice, this means that if the direction of an edge is reversed, all three basis vectors change  signs—ensuring antisymmetry in all subsequent projections and derived interactions.}

\textcolor{black}{In the scalarization step (Fig.~\ref{fig:plcons}–2), node vector features—such as velocity and angular velocity—are projected onto these edge-local frames, yielding scalar components. These projected scalars are then combined with other scalar node features to create  edge embeddings that are invariant to node ordering. This approach encodes both the directional and scalar information about local interactions, all while preserving the system’s underlying symmetries.}

\textcolor{black}{During vectorization (Fig.~\ref{fig:plcons}–3), the edge embeddings are decoded into physically meaningful interaction terms. First, internal forces are predicted as antisymmetric vectors \( \vec{F}_{ij} = -\vec{F}_{ji} \), representing changes in linear momentum per node and ensuring local conservation (Fig.~\ref{fig:plcons}–3a). To achieve this, three scalar coefficients  are extracted  from each edge embedding using learned functions. These coefficients  modulate the basis vectors of the edge-local reference frame, reconstructing the 3D force vector. Because  the edge embeddings are invariant under node interchange, the decoded scalar coefficients also satisfy \( f_{1,ij} = f_{1,ji} \), \( f_{2,ij} = f_{2,ji} \), and \( f_{3,ij} = f_{3,ji} \). Together  with the antisymmetric flipping  of the local basis vectors, this ensures that the reconstructed force vectors for edges \( i \rightarrow j \) and \( j \rightarrow i \) are equal in magnitude and opposite in direction, ensuring  \( \vec{F}_{ij} = -\vec{F}_{ji} \). This decoding mechanism directly embeds conservation of linear momentum into the architecture. Moreover, because the force vectors are constructed from  reference frame that is SO(3)-equivariant and T(3)-invariant, they inherit these symmetry properties.}

\textcolor{black}{Second, angular momentum changes are decoded as antisymmetric vectors \( \vec{A}_{ij} = -\vec{A}_{ji} \), following the same approach: scalar coefficients from the edge embedding are used to scale the local basis vectors (Fig.~\ref{fig:plcons}–3b). The resulting vectors represent the total angular momentum exchange between nodes \( i \) and \( j \), combining both spin and orbital components. Only the spin component, however, directly affects angular velocity. To isolate it, the orbital contribution—computed as the cross product of the predicted internal force and its lever arm—is subtracted from the total angular momentum change. This step relies on decoding a consistent point of force  application, \( \vec{x}_{0ij} = \vec{x}_{0ji} \), which is shared between  both directions of an edge between two interacting bodies (Fig.~\ref{fig:plcons}–3c). This reference point serves as the effective location where  internal forces act, enabling spin torque to be computed as  
\(
\boldsymbol{\tau}_{ij} = \vec{A}_{ij} - (\vec{r}_i - \vec{x}_0) \times \vec{F}_{ij}.
\)}

\textcolor{black}{While angular momentum is typically conserved globally about a fixed reference point, \textsc{Dynami-CAL GraphNet} instead enforces conservation locally at each edge by anchoring interactions to the predicted force application point. This localized formulation enables modular, fine-grained modeling of complex systems, scales efficiently to large graphs, and naturally incorporates non-central and dissipative effects. As demonstrated  in Supplementary Section~\S5.1, this edge-level formulation provably ensures global conservation of angular momentum under symmetry-preserving aggregation.  This follows directly from the antisymmetry of internal forces and the consistent, shared structure of the reference points used for each edge.}

\begin{figure}[h!]  
    \centering  
    \includegraphics[width=0.95\textwidth]{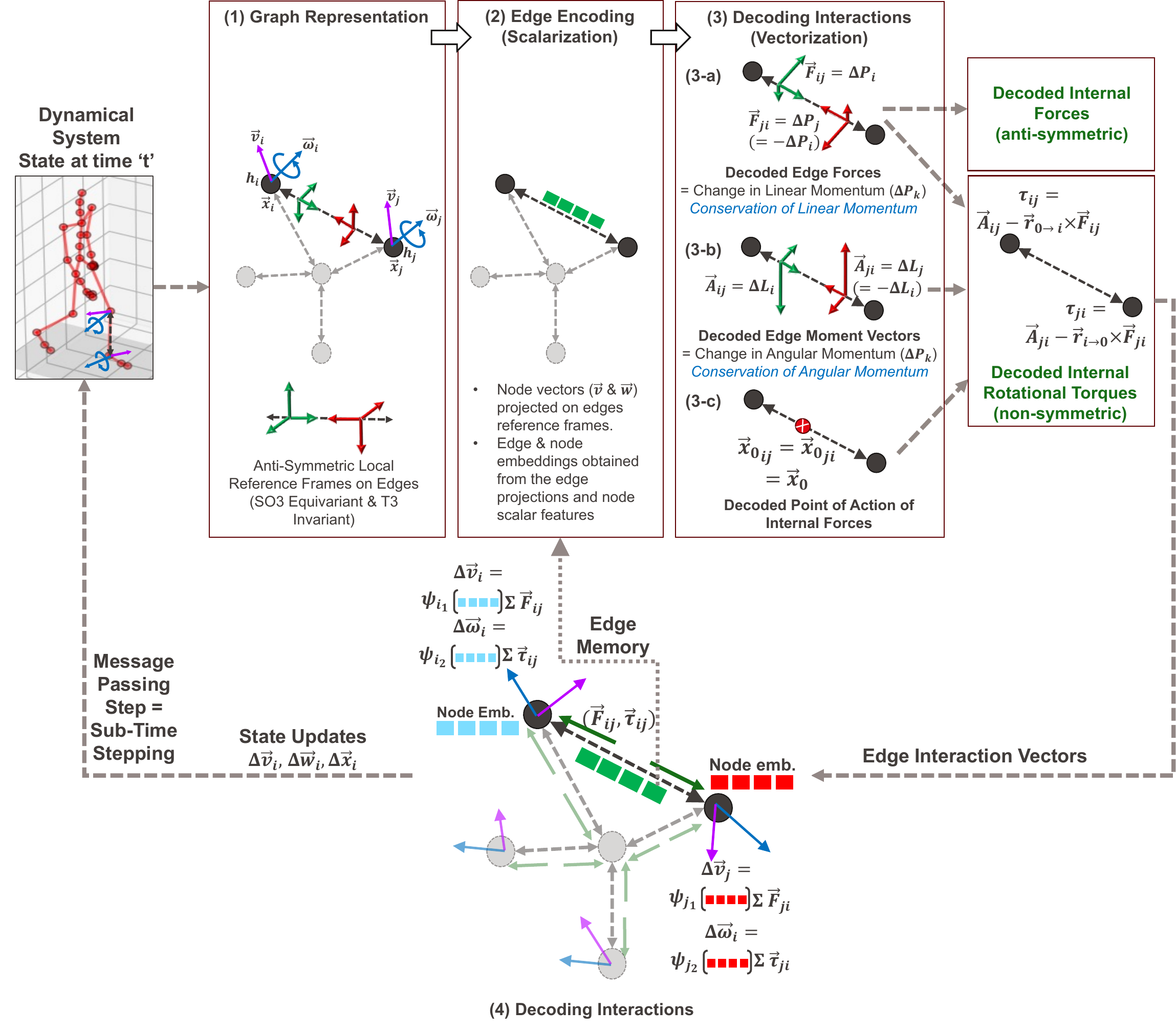}  
    \caption{\textcolor{black}{\textbf{Message Passing in \textsc{Dynami-CAL GraphNet}.}  
Decoded edge-wise internal force and torque vectors are aggregated at their respective  nodes and scaled using learned coefficients to update linear and angular velocities. These updates are then integrated using implicit Euler stepping to compute the new node positions. The edge embeddings are retained  as latent memory and used as skip connections to inform the edge embeddings in the subsequent  message-passing step. This mechanism enables the model to perform spatiotemporal reasoning within a single prediction step.
}}  
\label{fig:spatiotemp}  
\end{figure}

\textcolor{black}{\textbf{Spatiotemporal message passing.}
After computing physically consistent internal forces and torques at each edge, \textsc{Dynami-CAL GraphNet} aggregates these quantities at the node level. As illustrated  in Figure~\ref{fig:spatiotemp}, the decoded edge-wise internal forces and rotational torques are summed to obtain the net force and torque acting on each node. These vectors are then scaled by coefficients derived from the scalar node embeddings, resulting in updates to each node’s linear and angular velocities. The updated positions (and, optionally, orientations) are then computed using implicit Euler integration. This process constitutes a single message-passing layer of \textsc{Dynami-CAL GraphNet}.}

\textcolor{black}{Crucially, this message-passing step is  iteratively repeated to emulate sub-time stepping within a single prediction interval. At each iteration, the most recent  node states and the previously computed edge embeddings  are used to  inform the next round of edge encoding. This evolving representation is maintained as a latent memory on each edge—referred to  as \textit{Edge Memory} in Figure~\ref{fig:spatiotemp}-(4). As a result, the model achieves  spatiotemporal reasoning by continually enriching edge embeddings with both spatial context (from neighboring nodes) and temporal coherence (through accumulated interaction history across  message passing steps that mimic explicit time stepping). This design allows \textsc{Dynami-CAL GraphNet} to capture  dynamic behavior  over multiple time scales while preserving physically grounded inductive biases at each step.}

\textbf{Novel Mesh-Free and Particle-Free Modeling of Wall Boundaries}
To achieve a comprehensive representation of dynamic systems, it is essential to accurately model interactions with boundaries such as walls, floors, and rigid enclosures. These boundaries are integral to system behavior--constraining motion in robotic systems, supporting the body and  generating ground reaction forces and confining particles in granular simulations. \textcolor{black}{Existing approaches typically represent boundaries with  dense meshes or collections of particles~\cite{pfaff2020learning, allen2022learning, allen2023graph}. While effective, these approaches introduce significant computational overhead and require special treatment of wall-body interactions, distinct from body-body interactions. 
An alternative is to represent  boundaries implicitly by embedding additional features--such as distance to the wall--for each component \cite{sanchez2020learning}. Although this method is more efficient, it struggles in scenarios with multiple or moving boundaries, since fixed distance features cannot distinguish overlapping constraints or adapt to dynamic, time‐varying surfaces.}

\textcolor{black}{In this work, we address these limitations by proposing a novel mesh-free boundary treatment for GNNs that unifies body-wall and body-body interactions within  a single framework. Our approach reflects all nodes in the system across the outward normal of each boundary to create  ghost nodes. These ghost nodes inherit the  scalar properties of the boundary (e.g., degrees of freedom) and vector features (e.g., velocity and angular velocity); for stationary walls, the vector features are set to zero. Edges are then established  between original nodes and their corresponding ghost nodes according to a distance threshold---ensuring that only nodes sufficiently close to the boundary are connected to their reflections. These edges naturally encode body--wall interactions, capturing  normal forces, tangential reactions, and frictional effects. Additionally, ghost  nodes can inherit the motion of moving or rotating walls, allowing for the accurate modeling of dynamic  boundary effects. During message passing, their states are overwritten at each step with the wall's prescribed values, ensuring that boundary dynamics are correctly enforced and learned by the network.}

\textcolor{black}{The reflective mechanism  treats the boundary as an intermediate point between a node and its ghost, mirroring the modeling of  body--body interactions through center-to-center vectors. Because reflections are performed along the wall normals, the resulting edge directions naturally align with the boundary surface normals, ensuring  accurate modeling of interactions occurring in those directions. This approach requires  only  basic geometric information---such as the normal vector and a point on the plane for flat walls, or the axis and radius for cylindrical enclosures---making it both simple to implement and broadly applicable. }

\textcolor{black}{While  the method introduces $(W{-}1) \times N_n$ additional nodes for $W$ boundaries and $N_n$ physical nodes, this overhead is fixed and remains significantly lower  than that of  particle- or mesh-based boundary representations. For example, modeling large floors with explicit particles or meshes can quickly become infeasible due to memory constraints \cite{allen2022learning}, whereas  our method maintains  a constant  number of ghost nodes(i.e., $N_n$ for a single floor), regardless of the boundary's size. Moreover, in GNNs, the primary  computational cost arises from edge operations rather than the number of nodes. Since edges are created  only for nodes close to  the boundary, the additional  computational overhead remains minimal. These features make the proposed approach both scalable and physically consistent, enabling efficient modeling of complex boundary interactions in a wide range of  dynamical systems. Further details can be found in Section~\ref{s_s_sec:wall_model}}

\section{Results}\label{sec:results}

\subsection*{Overview of Experiments}

\textcolor{black}{We evaluate \textsc{Dynami‐CAL GraphNet} across four benchmarks spanning simulated and real-world physical systems. These include two simulated domains--(i) granular six-degree-of-freedom (6-DoF) collisions and (ii) charged particles connected by sticks or hinges—and two real-world domains characterized by complex spatiotemporal dynamics--(iii) human walking kinematics from Carnegie Mellon University (CMU) motion-capture data~\cite{cmu_mocap}, and (iv) protein molecular dynamics in water and ion solution under isothermal-isobaric (NPT: constant Number of particles, Pressure, and Temperature) conditions (300 K, 1 bar) \cite{seyler2017molecular}. These tasks capture  key challenges in modeling dynamical systems,  such as rotational dynamics, holonomic constraints, spatiotemporal coherence, and fine-scale conformational changes.  We compare \textsc{Dynami‐CAL GraphNet} with  several state-of-the-art baselines for each benchmark. All models are trained using single-step supervision, with multi-step rollouts used where applicable to assess long-horizon prediction accuracy and stability.
Overall, these evaluations demonstrate the ability of \textsc{Dynami‐CAL GraphNet} to model diverse dynamical systems with varying physical constraints, interaction types, and temporal scales.}

\subsection{\textbf{Granular 6-DoF collisions}} 
\label{s_sec:6dof_dem}
\textcolor{black}{This benchmark serves as a primary testbed for evaluating the ability of \textsc{Dynami‐CAL GraphNet} to model coupled translational and rotational dynamics under contact-rich, dissipative conditions. The dataset consists of 6-DoF trajectories of granular spheres undergoing inelastic collisions -- both inter-sphere and with enclosure walls -- simulated using the MFiX Discrete Element Method (DEM)~\cite{lu2022gpu,garg2012documentation,garg2012open}. The underlying physics includes non-linear normal and tangential contact forces, damping, Coulomb friction, and externally applied forces. All simulation details, parameters, and implementation setup are provided in Supplementary Information Section~\S1.1.}

\textcolor{black}{We evaluate model performance using  three physically grounded experiments that assess generalization, conservation behavior, and robustness to external forcing:}
\textcolor{black}{
\begin{enumerate}
    \item \textbf{Confined Granular collisions:} We introduce a  benchmark comprising five simulated trajectories of 60 identical spheres confined within a stationary cuboidal enclosure and initialized with random velocities. This setting evaluates the model’s ability to generate stable long-horizon rollouts and to capture physically consistent evolution of system-level kinetic energy, linear momentum, and angular momentum in an open, dissipative system -- where energy and momentum are absorbed at the stationary walls. Performance is assessed under both within-distribution (interpolation) and out-of-distribution (extrapolation) initial velocities. Dataset configuration, including training, validation, and test splits, learning objectives, and evaluation metrics are detailed in Supplementary Information Section~\S1.1. 
    \item \textbf{Oblique collision conservation:} The model trained on the homogeneous confined collision task is evaluated on a controlled two-sphere setup undergoing an oblique collision in a closed system. This benchmark assesses the model’s ability to conserve total linear and angular momentum in the absence of external forces, and provides an interpretable measure  of physical consistency.
    \item \textbf{Extrapolation to moving boundaries:} This task evaluates the model’s ability to generalize and extrapolate to previously unseen boundary conditions and large-scale system configurations. The model is trained on five trajectories of 60 spheres within a stationary cuboidal enclosure, where the spheres are influenced by gravity and interact via heterogeneous sphere–sphere and sphere–wall contact parameters (e.g., coefficients of restitution, friction angles, and stiffness values). At test time, the model is evaluated on a significantly more complex, real-world-inspired scenario: a rotating cylindrical hopper mixer  with curved walls, containing 2073 spheres and subjected to non-uniform rotational acceleration. Dataset details and the extrapolation test design are provided in Supplementary Information Section~\S1.2.1.
\end{enumerate}
}
\subsubsection{Confined Granular Collisions}
\label{sec:conf_gran_col}
\begin{figure}[h!]  
    \centering  
    \includegraphics[width=0.85\textwidth]{figures_results/interp_extra.pdf}  
\caption{\textcolor{black}{
\textbf{Long-horizon rollouts for confined granular collisions.}  
6-DoF rollouts of 60 spheres inside a cuboidal box under two regimes: interpolation (1a–1g) and extrapolation (2a–2g), with the latter tripling the initial kinetic energy relative to training.
\textbf{Interpolation (1a–1d):} System-level metrics—(a) number of spheres retained, (b) kinetic energy per unit mass, (c) largest component of linear momentum, and (d) largest component of angular momentum—tracked over 500 time steps. \textsc{Dynami-CAL GraphNet} retains all particles and accurately captures energy dissipation and momentum evolution, while GNS diverges early in all metrics.
\textbf{Extrapolation (2a–2d):} \textsc{Dynami-CAL GraphNet} remains stable and physically consistent under unseen initial velocities, whereas GNS fails to confine particles and exhibits large errors (metrics are calculated only for retained spheres).
\textbf{Rollout Snapshots (1e–1g, 2e–2g):} Selected time steps visualize spatial dynamics across Ground Truth, GNS, and \textsc{Dynami-CAL GraphNet}, highlighting the close match between predicted  \textsc{Dynami-CAL GraphNet} trajectories and ground truth.
}}

\label{fig:result_int_ext_coll}  
\end{figure}

\textcolor{black}{We evaluate the long-horizon rollout performance of \textsc{Dynami-CAL GraphNet} on a granular system of 60 identical spheres confined in a stationary cuboidal box. The training set comprises five DEM-simulated trajectories under zero gravity (see Supplementary Information Section~\S1.1.1 for dataset details). The model is evaluated in both within-distribution (interpolation) and out-of-distribution (extrapolation) regimes, with the latter initialized at approximately three times the kinetic energy observed  during training. In this open, dissipative system, energy is lost through inelastic collisions and momentum is absorbed by  the walls, causing the system to gradually settle over time.}

\textcolor{black}{We monitor  the evolution of kinetic energy, linear momentum, and angular momentum of the retained spheres over time, using metric formulations detailed  in Supplementary Information Section~\S1.1.2. Unlike existing dynamical systems benchmarks that primarily focus  on qualitative behavior or position accuracy,  our evaluation focuses on physically consistent rollouts and accurate learning of contact interactions -- both of which are critical for stable long-horizon prediction.}

\textcolor{black}{At each time step, the system is represented as a graph, where nodes correspond to spheres and encode  positional and dynamical features, including linear and angular velocities at times $t$ and $t{-}1$. Wall interactions are modeled via ghost nodes, which are reflections about the enclosure walls. The ghost nodes inherit the same properties as the boundaries--specifically in this case zero velocity and boundary identifiers. Edges are established  between  all sphere and ghost nodes  based on a distance threshold. All features are normalized  by their maximum values observed during training  to preserve directionality. The model is trained to predict per-sphere updates in position, linear velocity, and angular velocity, using the ground-truth differences between consecutive time steps as targets. These predicted updates are then integrated autoregressively during inference. Full implementation details are provided in Supplementary Section~\S1.1.2.}

\begin{figure}[h]  
    \centering  
    \includegraphics[width=0.75\textwidth]{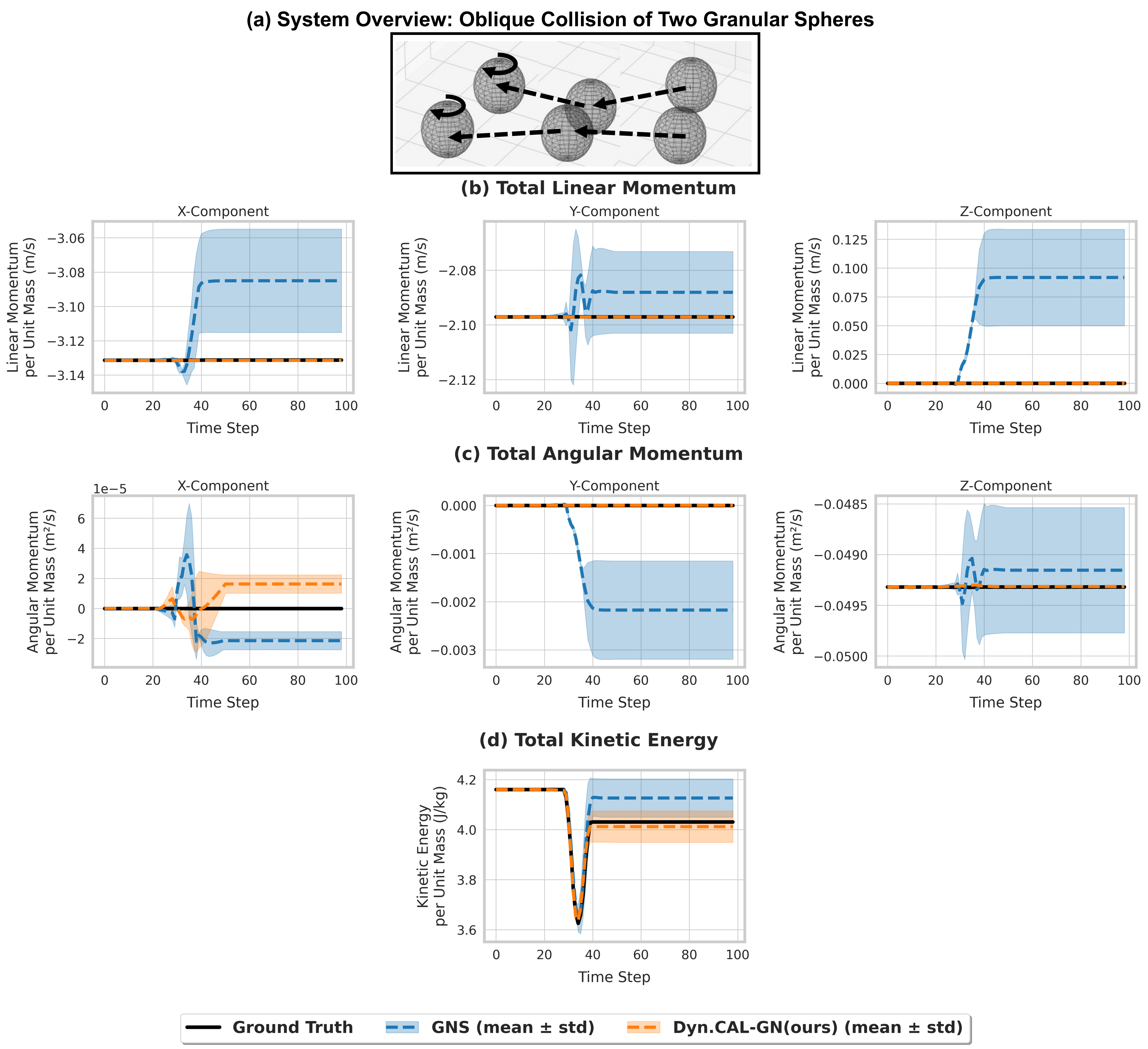}
\caption{\textcolor{black}{
\textbf{Oblique collision of two granular spheres.}  
(a) Collision trajectory.  
(b) Total linear momentum per unit mass.  
(c) Total angular momentum.  
(d) Total kinetic energy.  
\textsc{Dynami-CAL GraphNet} accurately conserves momentum and predicts energy dissipation, while GNS violates conservation laws.
}}
\label{fig:result_oblique_collision}  
\end{figure}

\textcolor{black}{We compare \textsc{Dynami-CAL GraphNet} with GNS~\cite{sanchez2020learning} and our reimplemented 6-DoF variants of EGNN~\cite{satorras2021n}, GMN~\cite{huang2022equivariant}, and ClofNet~\cite{du2022se}, where we extend the original architectures to predict angular velocity updates in addition to linear velocity and position updates at each time step. All models receive identical inputs, share the same training objectives, and utilize a common  reflection-based wall modeling approach (see Supplementary Information Section~\S1.1.3). The effectiveness  of our proposed boundary modeling strategy is further demonstrated in Supplementary Section~\S1.1.6, where GNS achieves  improved performance over its original distance-feature formulation. Among the  reimplemented baselines, EGNN, GMN, and ClofNet consistently underperform. Evaluated on 500-step rollouts in both interpolation and extrapolation settings, they exhibit clear deviations in kinetic energy decay as well as in the evolution of linear and angular momentum over time (see Supplementary Information Section~\S1.1.4 for detailed results). Their equivariant architectures struggle to model the non-linear, event-driven nature of inelastic collisions, while GNS—despite lacking equivariance—proves more expressive in capturing impulse-driven dynamics. As a result, GNS  is retained as the primary baseline in the main paper.}

\textcolor{black}{Figure~\ref{fig:result_int_ext_coll} compares \textsc{Dynami-CAL GraphNet} and GNS in both interpolation and extrapolation regimes. \textsc{Dynami-CAL GraphNet} reliably  retains all particles, accurately tracks kinetic energy decay, and preserves momentum evolution over 500 steps, exhibiting  low variance across random seeds.  In contrast, GNS diverges early  in the extrapolation setting, leading to particle escape 
due to its inability to generalize learned interactions under high-momentum conditions, where increased collision speeds require accurate resolution of impulsive contact forces to maintain confinement. Even for retained spheres it predicts increasing  deviations from expected physical behavior.} 

\textcolor{black}{These results highlight  the robustness and superior generalization ability  of \textsc{Dynami-CAL GraphNet} for  modeling dissipative, contact-rich 6-DoF dynamics.}

\subsubsection{Oblique Collision: Conservation of Linear and Angular Momentum}
\label{sec:two_body_ob_col}

\textcolor{black}{To assess  conservation behavior, we evaluate  models trained in the homogeneous confined setting (Section~\ref{sec:conf_gran_col}) using  a controlled two-sphere system undergoing an oblique, inelastic collision. Both spheres are initialized with zero angular velocity and assigned velocities to induce an angled impact. In this closed system, total linear and angular momentum should be conserved, while kinetic energy dissipates due to inelasticity.}

\textcolor{black}{Figure~\ref{fig:result_oblique_collision} presents results across three random seeds. \textsc{Dynami-CAL GraphNet} accurately preserves all components of linear and angular momentum and closely tracks the expected decay in kinetic energy. In contrast, the GNS baseline violates conservation laws and exhibits unphysical kinetic energy gain under one of the seeds. Supplementary Information Section~\S1.1.5, Figure 2, extends this comparison  to additional  baselines (EGNN, GMN, and ClofNet), which display over-damped behavior and fail to conserve linear and angular momentum.}

\textcolor{black}{These findings demonstrate  \textsc{Dynami-CAL GraphNet}'s ability to faithfully model contact-driven, multi-body dynamics with physically consistent impulse responses. Supplementary Figure 3 further quantifies post-collision errors, confirming \textsc{Dynami-CAL GraphNet}'s superior accuracy in capturing both translational and rotational dynamics.}

\subsubsection{\textcolor{black}{Extrapolation to Moving Boundaries: Rotating Cylindrical Hopper}}
\begin{figure}[h!]  
    \centering  
    \includegraphics[width=0.75\textwidth]{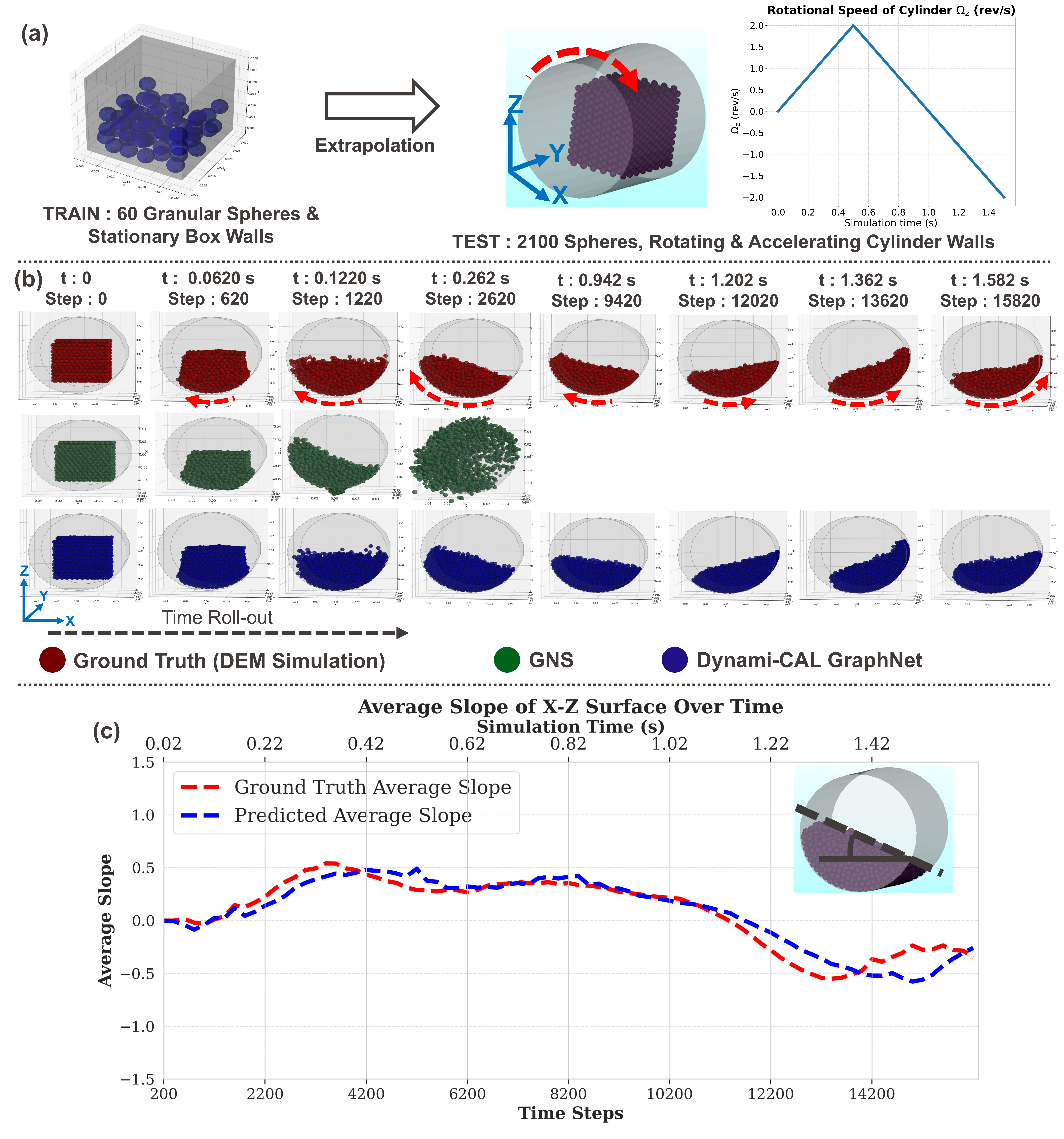}  
\caption{\textcolor{black}{\textbf{Extrapolation to rotating curved boundaries: performance in cylindrical hopper.}  
\textbf{(a)} \textsc{Dynami-CAL GraphNet}, trained on 60-sphere trajectories within stationary box walls, is evaluated on a cylindrical hopper with 2073 spheres and rotating walls. The wall rotation profile (rev/s vs. time) is shown at top right.  
\textbf{(b)} Rollout snapshots over 16,000 steps show that \textsc{Dynami-CAL GraphNet} (blue) accurately predicts particle motion and surface evolution, matching the DEM ground truth (red). In contrast, GNS (green) destabilizes early, failing to capture stable sphere–wall interactions.  
\textbf{(c)} Evolution of the average X–Z surface slope over time. The slope responds to changing rotation direction and speed; \textsc{Dynami-CAL GraphNet} accurately reproduces the ground truth surface evolution.
}}
\label{fig:result_cylinder_hopper}  
\end{figure}

\textcolor{black}{To assess generalization and extrapolation under complex external forcing and boundary geometries, we evaluate  \textsc{Dynami-CAL GraphNet} on a real-world-inspired granular mixing task: 2,073 spheres in a rotating cylindrical hopper -- a scenario  relevant to industrial mixing and particulate flow applications. The model is trained exclusively  on five trajectories, each consisting of 1,500 time steps, involving 60 spheres confined within a stationary cuboidal box and subject to gravity, featuring heterogeneous sphere–sphere and sphere–wall contact parameters. During testing, particle reflections are computed dynamically, and the corresponding interaction graph is created at each rollout step based on the curved hopper walls and their instantaneous motion (see Supplementary Information Section~\S1.2.1 for dataset details, and Sections~\S1.2.2 and~\S1.2.3 for implementation of \textsc{Dynami-CAL GraphNet} and GNS with boundary adaptation). Figure~\ref{fig:result_cylinder_hopper}(a) illustrates the training and test configurations.} 

\textcolor{black}{At test time, the hopper rotates about the Y-axis with a time-varying angular velocity, generating tangential impulses that induce a dynamic surface slope in the X–Z plane. Remarkably, despite being trained only on flat, stationary boundaries, \textsc{Dynami-CAL GraphNet} delivers highly accurate  predictions over 16000-rollout steps. The model not only tracks the detailed spatial trajectories  of thousands of particles (Fig.\ref{fig:result_cylinder_hopper} (b)) , but also precisely captures the evolving macroscopic surface slope  throughout the entire simulation (Fig.\ref{fig:result_cylinder_hopper} (c)).} 


\textcolor{black}{In contrast, the GNS baseline (see implementation details for this case in Supplementary Section~\S1.2.3) destabilizes early and fails to generalize to the novel boundary conditions. These results highlight the strong extrapolation capability of \textsc{Dynami-CAL GraphNet}, demonstrating generalization across configurations, initial conditions, and boundary regimes. \textbf{This experiment further underscores \textsc{Dynami-CAL GraphNet}'s flexibility as a deployable, general-purpose simulator -- capable of handling diverse geometries and evolving environments without retraining to each new configuration, architectural changes or ad hoc interventions such as remeshing.}}

\vspace{1mm}

\textcolor{black}{Additional evaluations presented  in the Supplementary Information further demonstrate the model’s robustness and versatility: it delivers accurate angular responses across  a range of impact angles (Section~\S1.1.7), maintains stability under sparse temporal sampling and stiff interactions (Section~\S1.1.8), and provides interpretable force decomposition into tangential and normal components (Section~\S1.1.9).}
  
\subsection{\textcolor{black}{\textbf{Constrained \texorpdfstring{$N$}{N}-Body Dynamics.}}}  
\label{s_sec:cons_nbody}
\textcolor{black}{To evaluate applicability to systems with mixed interaction types and structural constraints, we use the Constrained N-Body dataset introduced in~\cite{huang2022equivariant}, which extends the 3D charged particle simulation of Kipf et al.~\cite{kipf2018neural} by incorporating holonomic constraints in the form of rigid sticks and hinges (see Supplementary Information Section~\S2.1 for dataset details). This benchmark presents a challenging testbed for dynamics prediction, as it combines long-range Coulomb interactions with constraint-induced couplings that govern collective motion. Given the state of the system at time $t$, the target is to predict the future state at time $t+10$, which corresponds to 1000 simulation time steps.}


\textcolor{black}{We benchmark  \textsc{Dynami-CAL GraphNet} against several strong baselines: GMN~\cite{huang2022equivariant} ( constraint enforcement  via generalized coordinates and handcrafted forward kinematics), EGNN~\cite{satorras2021n} (lightweight E(\(n\))-equivariant message passing), EGNNReg~\cite{huang2022equivariant} ( explicit constraint penalties), Radial Field Networks (RF)~\cite{kohler2019equivariant} (E(\(n\))-equivariant updates based on edge distances), Tensor Field Networks (TFN)~\cite{thomas2018tensor} (SE(3)-equivariant feature propagation with  spherical harmonics), SE(3)-Transformer~\cite{fuchs2020se} (attention-based extension of TFN), ClofNet~\cite{du2022se} (edge-wise local reference frames), a message-passing GNN \cite{gilmer2017neural}, and a linear kinematic predictor  \(p(t) = p(0) + v(0)t\). } 

\textcolor{black}{For fair comparison, we adopt the training and evaluation settings presented in~\cite{huang2022equivariant} for configuring \textsc{Dynami-CAL GraphNet} (see Supplementary Information Section~\S2.2 for implementation details). For EGNN, EGNNReg, RF, TFN, SE(3)-Transformer, and the standard message-passing GNN, we report results directly from~\cite{huang2022equivariant}, all obtained under the same experimental setup. In our experiments, we additionally evaluate ClofNet on this benchmark using its publicly released implementation, aligning its configuration with that of the other baselines. We also conduct additional rollout evaluations for GMN using its publicly available code and default settings. Further details on all baseline models are provided in Supplementary Information Section~\S2.3. All models are trained using single-step supervision; for multi-step rollout evaluation, we report results for GMN only, which is the strongest-performing one-step baseline. For rollout, both GMN and \textsc{Dynami-CAL GraphNet} were trained to predict the single-step position and velocity targets. The original paper~\cite{huang2022equivariant} did not evaluate GMN under multi-step rollout; we include this to assess long-term stability and physical consistency in predicted dynamics.}

\begin{figure}[h!]  
    \centering  
    \includegraphics[width=1.0\textwidth]{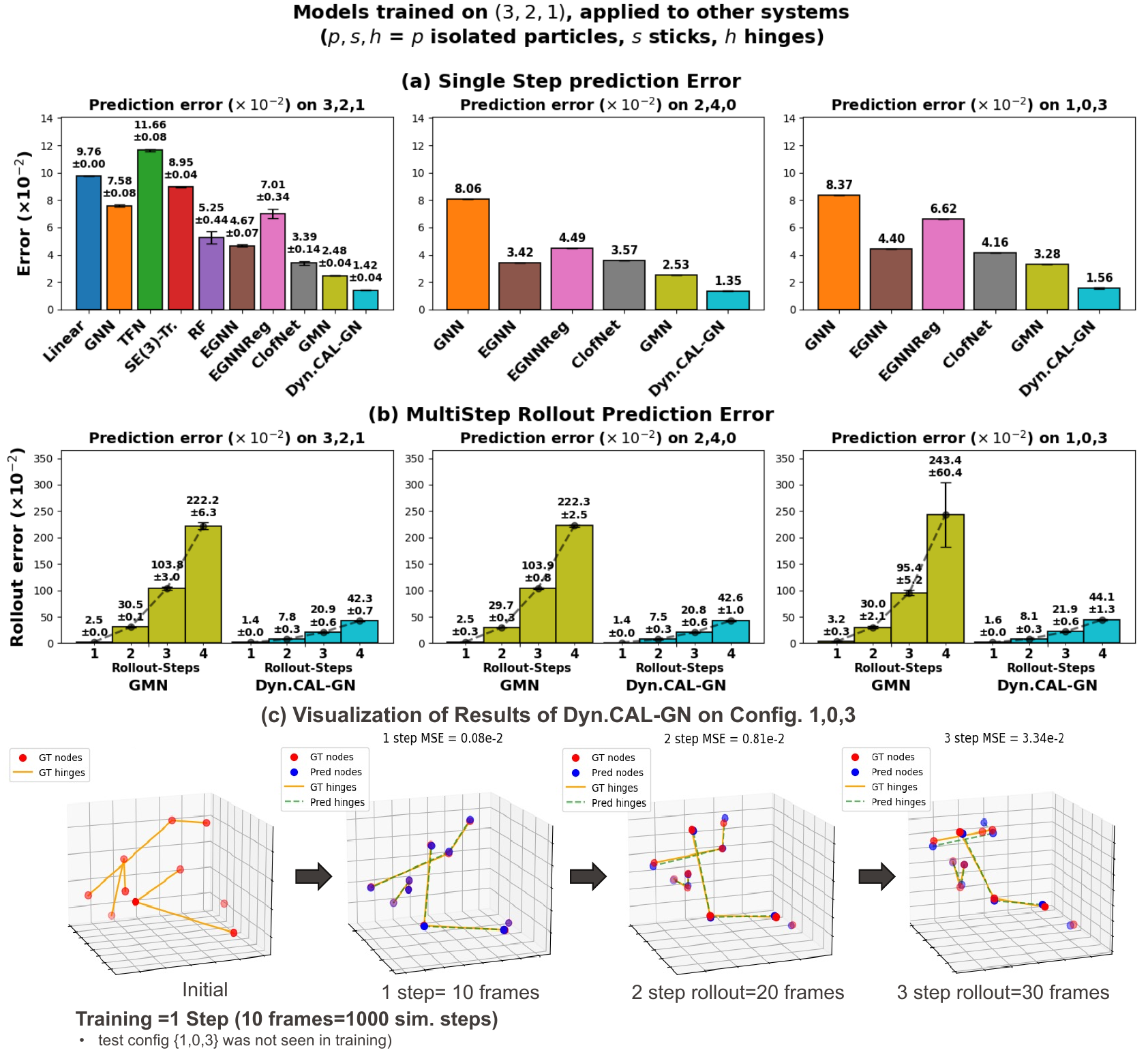}  
\caption{
\textcolor{black}{\textbf{Performance on the Constrained \(N\)-Body benchmark.}  
Models are trained on the (3,2,1) configuration and evaluated on both seen and unseen systems: (3,2,1), (2,4,0), and (1,0,3).  
(a) \textbf{Single-step error:} \textsc{Dynami-CAL GraphNet} achieves the lowest error across all configurations.  
(b) \textbf{Multi-step rollout:} Compared to GMN, our model maintains stable accuracy over long horizons.  
(c) \textbf{Qualitative rollout:} Accurate constrained dynamics on unseen (1,0,3) system.
}}
\label{fig:result_nbody}  
\end{figure}

\textcolor{black}{Figure~\ref{fig:result_nbody} summarizes the results  on the Constrained \(N\)-Body benchmark. \textsc{Dynami-CAL GraphNet} consistently outperforms all baseline models  in both single- and multi-step prediction tasks. In Figure~\ref{fig:result_nbody}(a), our model achieves the lowest single-step prediction error on both  seen (3,2,1) and unseen (2,4,0) and (1,0,3) configurations, surpassing GMN, EGNN, and ClofNet. As shown in Figure~\ref{fig:result_nbody}(b), \textsc{Dynami-CAL GraphNet} maintains stable long-horizon accuracy  over multi-step rollout up to four steps (one step = 10 frames = 1,000 simulation steps), whereas  GMN accumulates significant error over time. Figure~\ref{fig:result_nbody}(c) presents qualitative rollouts on the unseen (1,0,3) configuration, demonstrating that our model accurately captures constrained dynamics, despite being trained only with single-step supervision on a different topology.}

\begin{figure}[h!]  
    \centering  
    \includegraphics[width=0.6\textwidth]{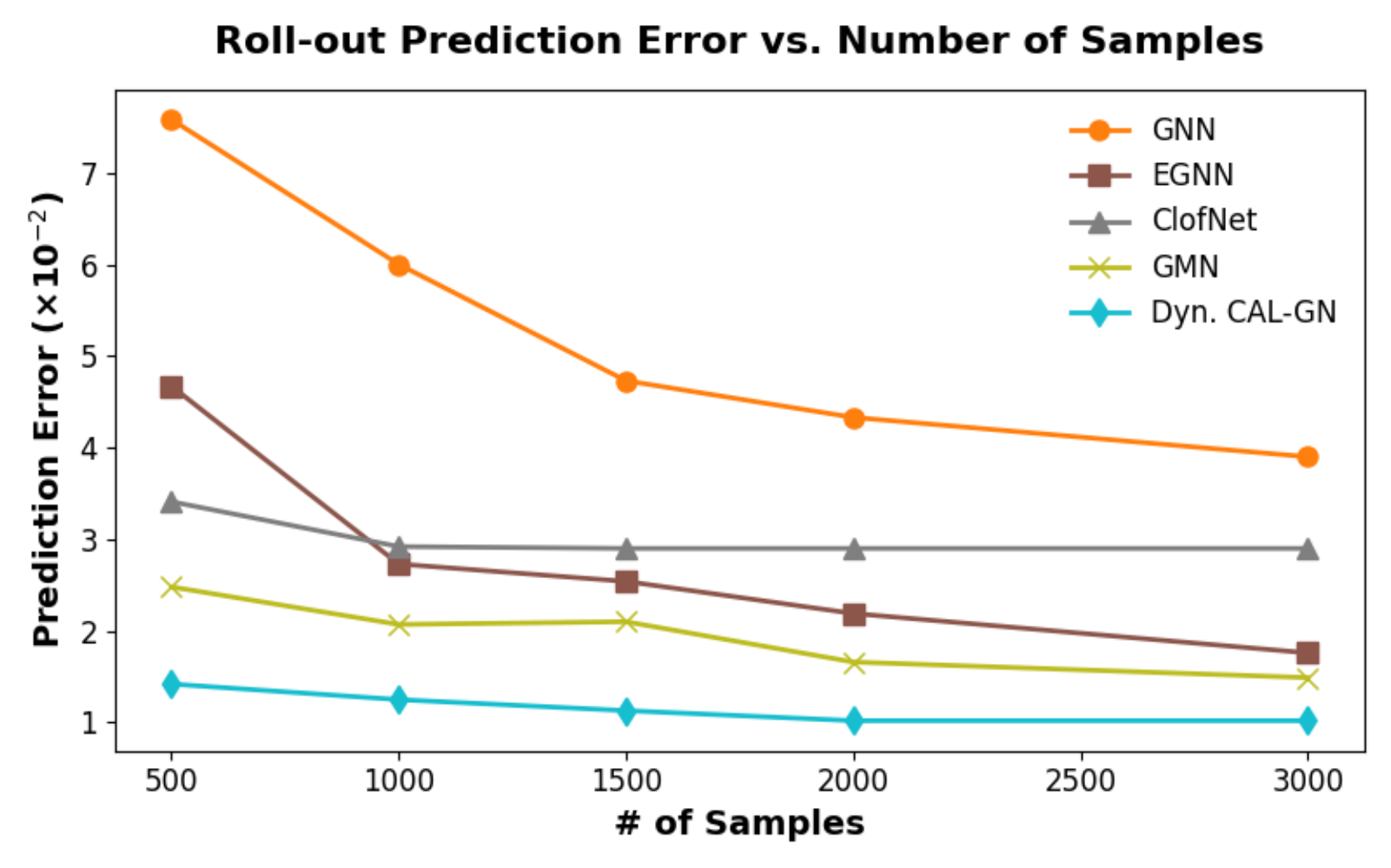}  
\caption{\textcolor{black}{\textbf{Rollout prediction error vs. number of training samples} on constrained \(N\)-body (3,2,1). \textsc{Dynami-CAL GraphNet} (blue) achieves the best performance across all training sizes, demonstrating strong data efficiency.}}
\label{fig:dat_eff}  
\end{figure}

\textcolor{black}{While all models receive edge-type labels (e.g., stick or hinge), \textsc{Dynami-CAL GraphNet} uniquely exploits  this information to infer physically consistent internal forces and moments, without relying on explicit constraint formulations, as  in GMN. GMN enforces structural constraints through  generalized coordinates and a handcrafted forward kinematics module, making it susceptible to integration errors and constraint drift over long rollouts. In contrast, \textsc{Dynami-CAL GraphNet} employs  evolving edge embeddings that serve as latent memory units -- conditioned on edge type and iteratively updated through message passing across spatial neighbors and multiple sub-time steps -- to effectively capture temporal dynamics. Coupled  with an architecture grounded in physical laws, such as conservation of linear and angular momentum, this approach enables the model to learn robust and generalizable dynamics across a wide range of constrained systems. The importance of these components is further supported by ablation results on the (3,2,1) setup (Supplementary Section~\S2.4), which show that removing either conservation laws or spatiotemporal message passing substantially impairs performance.}

\paragraph{\textcolor{black}{Data Efficiency.}}
\textcolor{black}{We assess  data efficiency on the constrained \(N\)-body (3,2,1) setup by varying the number of training samples. This dataset is chosen to enable direct benchmarking against the reported data efficiency results in~\cite{huang2022equivariant} which include GNN, EGNN, and GMN. In addition, we evaluate ClofNet using its publicly available implementation under the same settings.}

\textcolor{black}{Figure~\ref{fig:dat_eff} presents single-step prediction errors for \textsc{Dynami-CAL GraphNet} and all considered baselines across different training set sizes.}

\textcolor{black}{\textsc{Dynami-CAL GraphNet} demonstrates  strong performance even with as few as 500 training samples, exhibiting  only marginal improvement  as more data is added. This highlights the model’s  ability to learn robust dynamics from limited data.} 

\subsection{\textcolor{black}{\textbf{Human motion prediction.}}}

\begin{figure}[h!]  
    \centering  
    \includegraphics[width=0.95\textwidth]{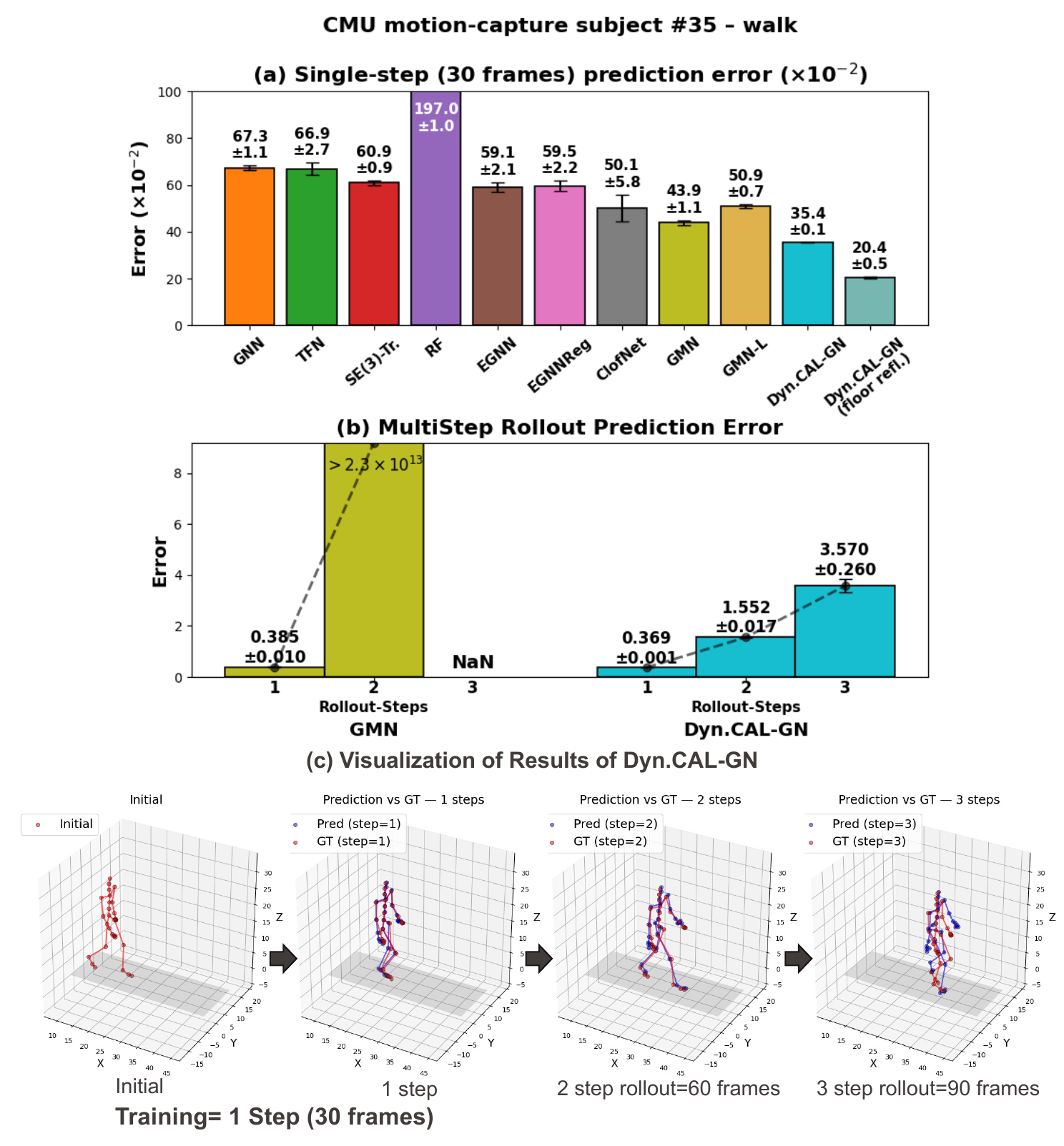}  
    \caption{\textcolor{black}{
    \textbf{Performance on the CMU Motion Capture benchmark (subject~\#35, walk).} 
    (a) \textbf{Single-step prediction error:} \textsc{Dynami-CAL GraphNet} achieves the lowest error among all baselines, with the floor-reflection variant performing best. 
    (b) \textbf{Multi-step rollout error:} While GMN diverges rapidly, \textsc{Dynami-CAL GraphNet} sustains low error over time. 
    (c) \textbf{Qualitative rollout:} Despite being trained with single-step supervision, the model produces stable predictions that accurately track the ground truth over three rollout steps (90 frames), demonstrating its ability to learn spatiotemporal dynamics effectively.
    }}
\label{fig:result_humanbody}  
\end{figure}

\textcolor{black}{We evaluate \textsc{Dynami-CAL GraphNet} on a real-world benchmark using the CMU Motion Capture dataset \cite{cmu_mocap}, chosen to assess the model’s ability to capture articulated, constrained dynamics from real-world motion data. The dataset records  articulated 3D joint trajectories during various human activities. Following the data split presented in~\cite{huang2022equivariant}, we use walking sequences from subject~\#35. Dataset splits and preprocessing are detailed in Supplementary Information Section~\S3.1. The task involves  predicting the positions and velocities of all joints at a future time step (\(t{+}30\)), given their current state at time \(t\). This requires the model to reason over both spatial articulation and temporal dynamics  using partial observations (i.e., joint motion only, without external ground reaction forces).}

\textcolor{black}{Implementation details of \textsc{Dynami-CAL GraphNet} are described in Supplementary Information Section~\S3.2. We further assess a variant, \textsc{Dynami-CAL GraphNet} (floor refl.), which augments the skeleton with ghost foot nodes by reflecting the original foot nodes across the ground plane -- defined as the minimum \(z\)-coordinate at each frame -- using the same reflection scheme as in the 6-DoF benchmark (Section~\ref{sec:conf_gran_col}). These ghost nodes are connected via 1-hop edges to the foot nodes and inherit ground features—i.e., zero velocity and a distinct ground label (2), distinguishing them from the foot nodes (labeled 1) and the rest of the joints (labeled 0)—enabling the model to better capture  ground contact behavior.}

\begin{figure}[h!]  
    \centering  
    \includegraphics[width=0.95\textwidth]{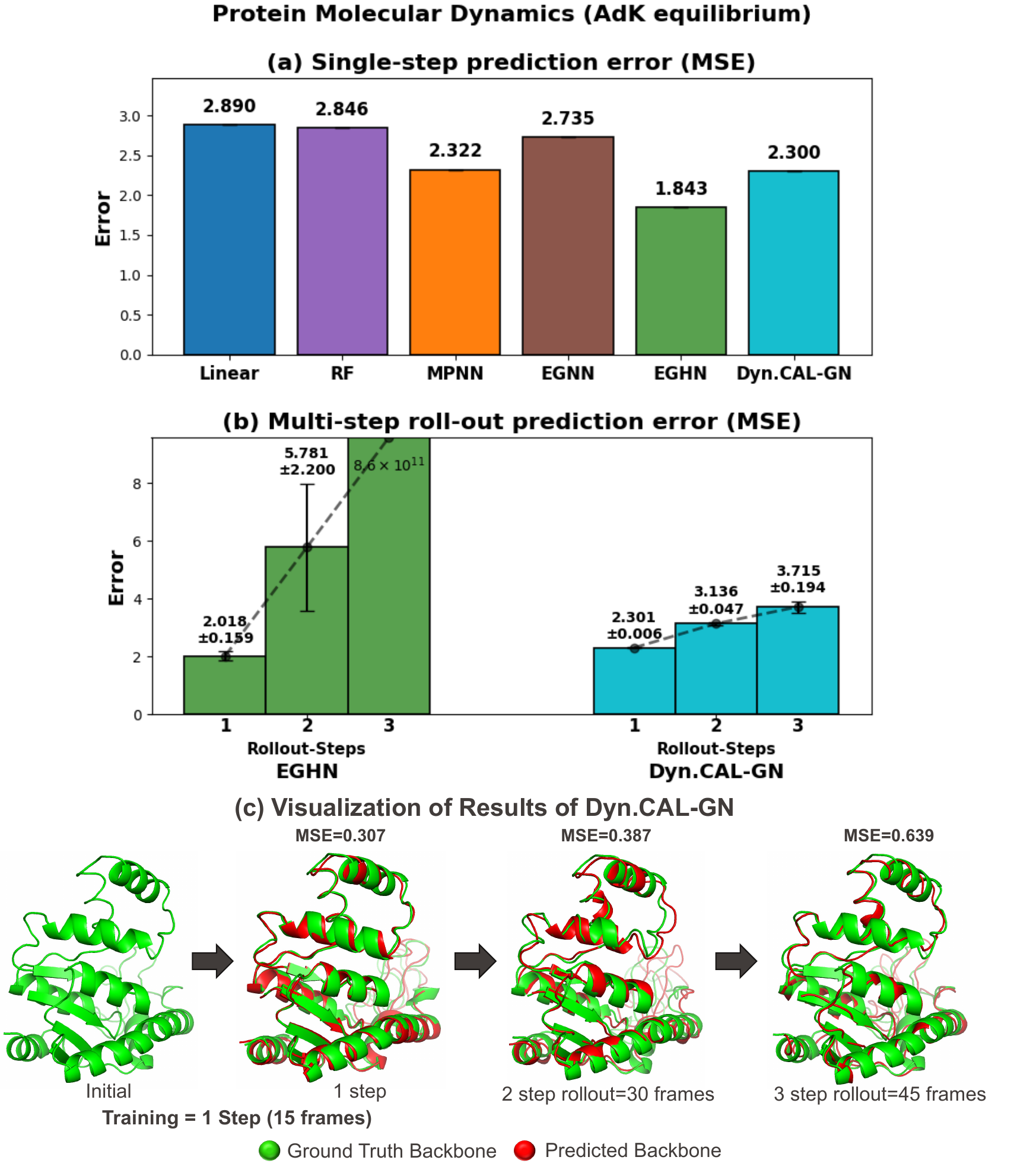}  
    \caption{\textcolor{black}{
    \textbf{Protein molecular dynamics (AdK equilibrium).} 
    (a) \textbf{Single-step prediction error:} \textsc{Dynami-CAL GraphNet} achieves the second-lowest mean squared error (MSE), closely trailing EGHN and outperforming MPNN, EGNN, RF, and a linear baseline. One prediction step corresponds to 15 trajectory frames.  
    (b) \textbf{Multi-step rollout error:} While EGHN diverges rapidly, \textsc{Dynami-CAL GraphNet} maintains stable and accurate multi-step rollout predictions.  
    (c) \textbf{Qualitative rollout:} Predicted backbone structures from \textsc{Dynami-CAL GraphNet} are shown starting from an initial protein configuration and compared to ground-truth conformations at each rollout step (up to 3 steps or 45 simulation frames). The model closely tracks conformational evolution despite training with only single-step supervision.
    }}
\label{fig:result_protein}  
\end{figure}

\textcolor{black}{We compare  our approach against a comprehensive set of baselines evaluated in prior work~\cite{huang2022equivariant} (see Supplementary Information Section~\S3.3), including models with handcrafted kinematics (GMN~\cite{huang2022equivariant}) and its learned variant (GMN-L), as well as a range of equivariant graph neural networks: EGNN~\cite{satorras2021n}, EGNNReg~\cite{huang2022equivariant}, TFN~\cite{thomas2018tensor}, SE(3)-Transformer~\cite{fuchs2020se}, Radial Field Networks (RF)~\cite{kohler2019equivariant}, ClofNet~\cite{du2022se}, and a message-passing GNN~\cite{gilmer2017neural}.}

\textcolor{black}{Figure~\ref{fig:result_humanbody}(a) reports  single-step prediction accuracy on the CMU human walk benchmark. \textsc{Dynami-CAL GraphNet} achieves the lowest error among  all compared methods, outperforming GMN, which models the human skeleton using 19 joints and 6 manually specified  rigid links (e.g., (0,11), (2,3), (7,8)) enforced by  a handcrafted forward kinematics (FK) module. Notably, the \textsc{Dynami-CAL GraphNet} (floor refl.) variant, which augments the skeleton with ghost foot nodes reflected across the ground plane, achieves even lower errors across three random seeds.}

\textcolor{black}{Figure~\ref{fig:result_humanbody}(b) demonstrates that, despite training with only single-step supervision, \textsc{Dynami-CAL GraphNet} maintains stable accuracy during multi-step rollouts, whereas GMN quickly diverges. Qualitative results in Figure~\ref{fig:result_humanbody}(c) further  confirm that predicted joint trajectories remain coherent and physically plausible, closely  tracking ground-truth motion over 90 future frames. In this experiment, both GMN and \textsc{Dynami-CAL GraphNet} were trained for 1000 epochs to predict the target position and velocity.}

\textcolor{black}{Together with the results on the constrained \(N\)-body system (Section~\ref{s_sec:cons_nbody}), these findings provide strong empirical evidence that \textsc{Dynami-CAL GraphNet} performs effective spatiotemporal reasoning by evolving edge embeddings over time. These embeddings are enriched with spatial context  through  message passing between  neighboring nodes and are temporally propagated via  iterative updates that mimic  sub-time stepping. Crucially, this mechanism is grounded in universal physical inductive biases -- namely,  conservation of linear and angular momentum governing internal forces and moments -- which apply  across diverse physical systems. As a result,  the model is able  to learn localized physical interactions over space and time, enabling robust and generalizable dynamics across a wide range of structured physical systems.}

\subsection{\textcolor{black}{\textbf{Protein dynamics in solvent.}}  }

\textcolor{black}{We choose this benchmark to assess \textsc{Dynami-CAL GraphNet}'s ability to capture fine-grained conformational changes across multiple spatial and temporal scales, and to demonstrate its applicability to a real-world scientific domain. In this benchmark, we evaluate its capacity to model complex, thermally driven protein dynamics that give rise to both global and local structural rearrangements. Specifically, the task involves predicting protein motion within a thermally fluctuating, high-dimensional molecular environment by forecasting the future positions of heavy atoms from their current configuration.} 

\textcolor{black}{We use the apo adenylate kinase (AdK) equilibrium trajectory dataset~\cite{seyler2017molecular}, accessed via the MDAnalysis toolkit~\cite{gowers2019mdanalysis}, which tracks  the atomistic motions of the protein solvated in explicit water and ions. This setup closely mirrors  a realistic aqueous cellular environment under near-physiological conditions (300~K, 1~bar), where the protein exhibits  both global conformational transitions and local side-chain rearrangements, driven by thermal fluctuations  and solvent interactions. The forecasting task --  predicting the future positions of heavy atoms -- presents a significant challenge  due to the  stochastic nature and structural flexibility of biomolecules. We choose this dataset for its biological relevance and physical complexity, and we follow the experimental setup and data-split protocol introduced \cite{han2022equivariant}. The task is to predict the protein's conformation at time $t+15$ given the state at time $t$. Further dataset details are provided in Section~\S2.1 of the Supplementary Information.}

\textcolor{black}{\textsc{Dynami-CAL GraphNet} represents the protein as a graph, where nodes correspond to backbone heavy atoms and are assigned their 3D positions, current velocities at time \(t\), and velocities from the previous step \(t{-}1\), computed via finite differencing of recorded trajectory positions during preprocessing. Interactions are modeled with edges that represent covalent bonds between atoms. Importantly, we restrict the edge structure to backbone covalent bonds only, as augmenting it based on geometric proximity (e.g., using a 10\,\AA{} cutoff) resulted in reduced performance. Implementation details are provided in Supplementary Information Section~\S2.2.}


\textcolor{black}{We compare our approach with EGHN (Equivariant Graph Hierarchical Network)~\cite{han2022equivariant}, a U-Net-style architecture that captures  both local and global molecular interactions while preserving geometric equivariance. EGHN  integrates  message passing  with hierarchical pooling and unpooling operations to model  fine-grained atomic details  as well as  broader structural patterns. Local edges correspond to covalent bonds, while global edges connect  atoms within a 10\,\AA{} distance threshold,  enabling the modeling of long-range interactions. In addition to EGHN, we  report results for several  baseline models from ~\cite{han2022equivariant}: Linear, EGNN~\cite{satorras2021n},Radial Field Networks (RF)~\cite{han2022equivariant}, and a message passing neural network originally proposed for molecular dynamics (MPNN)~\cite{gilmer2017neural}.}

\textcolor{black}{Figure~\ref{fig:result_protein}-(a) presents single-step prediction errors (mean squared error, MSE). \textsc{Dynami-CAL GraphNet} achieves the second-best performance, closely following  EGHN, and surpassing EGNN, RF, and the linear baseline. Although MPNN shows  competitive MSE, it lacks rotational equivariance and is highly sensitive to test-time transformations: as demonstrated  in~\cite{han2022equivariant}, applying a random rotation during evaluation increases its MSE dramatically  to 605.7.}

 \textcolor{black}{Figure~\ref{fig:result_protein}-(b) evaluates multi-step rollouts -- again, for models trained only with single-step supervision. While EGHN diverges quickly beyond the second step, \textsc{Dynami-CAL GraphNet} maintains accurate predictions for up to 3 steps corresponding to 45 frames of simulated trajectory. This temporal stability is further illustrated in Figure~\ref{fig:result_protein}-(c), where overlays of  predicted and ground-truth protein backbones show that our model's predictions remain conformationally faithful over time, capturing both large-scale structure and fine details. Although multi-step rollouts were not part of the original EGHN evaluation in \cite{han2022equivariant}, we apply them here to both EGHN and \textsc{Dynami-CAL GraphNet} as a stringent test of physical fidelity—where sustained stability under autoregressive prediction indicates fidelity of the learned dynamics.}

\textcolor{black}{These results highlight  the exceptional capability  of \textsc{Dynami-CAL GraphNet} to model physically consistent dynamics in complex, fine-grained systems such as proteins. This  accuracy is rooted in two core principles: (i) modeling  internal forces and moments by enforcing  conservation of linear and angular momentum, and (ii) capturing the spatiotemporal evolution of edge embeddings -- enriched with spatial context from neighboring nodes and  propagated through  iterative message passing that emulates sub-time stepping. }

\section{\textcolor{black}{Discussion}}
\textcolor{black}{In this work, we propose \textsc{Dynami-CAL GraphNet}, a physics-informed graph neural network to model six-degree-of-freedom dynamics in diverse multi-body systems. The architecture embeds conservation of linear and angular momentum as an inductive bias, enabling the model to learn physically consistent interactions—including those involving external forces and dissipation—directly from data. By employing  edge-local reference frames that are equivariant to rotations, invariant to translations, and antisymmetric under node interchange, the model  captures both geometric symmetries and fundamental conservation laws. Interactions are aggregated and integrated through  a multi-step message-passing scheme that mimics sub-time stepping, allowing for stable long-horizon predictions  and interpretable dynamics. As a result, \textsc{Dynami-CAL GraphNet} provides  a general, scalable, and physically principled  framework for learning dynamics in systems ranging from granular materials and  molecular assemblies to human motion.}

\textcolor{black}{We demonstrated the versatility  of \textsc{Dynami-CAL GraphNet} across four diverse benchmarks, encompassing both simulated and real-world systems. On the 6-DoF granular benchmark introduced in this work—characterized by dissipation and external forcing—the model learned physically consistent dynamics from just five training trajectories involving  60 spheres and successfully extrapolated to a rotating hopper with over 2,000 particles, maintaining stable rollouts over 16,000 time steps. In the constrained $N$-body benchmark, \textsc{Dynami-CAL GraphNet} outperformed baselines by learning holonomic constraints and enabling stable multi-step rollouts, even on unseen configurations. Applied to real-world human motion capture, the model inferred joint dynamics directly from data and produced stable multi-step predictions, surpassing constraint-aware baselines. In the protein dynamics benchmark, it captured  both fine-scale fluctuations and large-scale conformational changes  over extended horizons, outperforming a hierarchical baseline specifically designed for this task. Collectively, these results highlight the robustness, generalization ability, and physical fidelity of \textsc{Dynami-CAL GraphNet} across a broad range of dynamical systems.} \textcolor{black}{These strengths also position \textsc{Dynami-CAL GraphNet} as an efficient surrogate in scenarios where traditional simulation pipelines are challenged by frequent reconfiguration or limited knowledge of underlying dynamics.}

\textcolor{black}{At its core, \textsc{Dynami-CAL GraphNet} internalizes fundamental conservation laws by treating each interaction as an instantaneous closed system, thereby guaranteeing  the preservation of  linear and angular momentum as well as  translational and rotational symmetries—directly reflecting  Noether’s theorem and Newton’s third law}.

A natural extension of this framework is to continuum mechanics, where conservation laws are equally essential. For example, in finite element analysis (FEA), the Cauchy stress relation enforces linear momentum balance, and the symmetry of the  stress tensor ensures angular momentum conservation;  in fluid dynamics, the Navier–Stokes equations encode momentum conservation. \textcolor{black}{Adapting \textsc{Dynami-CAL GraphNet} to such domains   introduces new challenges, especially in enforcing local conservation laws across fields with effectively infinite degrees of freedom.}

\textcolor{black}{An additional challenge arises  in partially observed or unclosed systems—such as when external agents, like  unmodeled magnetic fields, act on  otherwise closed dynamics—where strict momentum conservation is no longer observed. Addressing these cases requires explicit modeling of external forces at the node level, conditioned on latent state embeddings. In our experiments, this was straightforward for gravitational forces, which depend only on scalar node embeddings (representing masses) and can be decoded independently at each node. However, modeling  more complex, context-dependent external forces remains an open direction  for future research.}

\textcolor{black}{By anchoring  learned dynamics in fundamental physical principles while retaining the flexibility to accommodate  real-world complexities, \textsc{Dynami-CAL GraphNet} provides a strong foundation for advancing data-driven modeling of physical systems. Extending this approach to continuum domains and partially observed environments opens exciting avenues for future research at the intersection of machine learning, physics, and engineering.}

\section{Method}
\label{sec:method}
\subsection{Graph representation of multi-body dynamical system}
\label{s_sec:graph_rep}


\textcolor{black}{We represent a multi-body dynamical system as a graph} \( G = (V, E) \), where \( V = \{v_i \mid i = 1, 2, \ldots, n\} \) denotes the set of bodies, and \( E = \{(e_{ij}, e_{ji}) \mid i \neq j, (i, j) \in V \times V\} \) represents bidirectional edges that encode interactions between distinct bodies, excluding self-loops. \textbf{ Edge connectivity} is determined either from the system's known geometry or dynamically computed using metrics such as Euclidean distance. \textcolor{black}{For example, in the granular collision dynamics examined in Section~\ref{s_sec:6dof_dem}, edges are formed between bodies \( i \) and \( j \) if \( \|\vec{r}_i - \vec{r}_j\| \leq d_c \), where \( d_c\) is the threshold distance criterion (taken as 1.25\(\times\)sphere diameter in the granular system). }


Each node \( v_i \) in the graph is assigned two types of \textbf{node features} in addition to the \textbf{position vectors}: \textbf{1)Vector Features:} \( \mathbf{V}_i = [\vec{v}_i^t, \vec{\omega}_i^t, \vec{v}_i^{t-1}, \vec{\omega}_i^{t-1}] \), which include the linear velocity \( \vec{v}_i^t \) and angular velocities  \( \vec{\omega}_i^t \) at the current time step $t$, as well as their values \( \vec{v}_i^{t-1} \) and \(\vec{\omega}_i^{t-1} \) at the previous time step $t-1$. 
\textcolor{black}{The \textbf{scalar features} \( \alpha_i \) represent categorical or continuous attributes that encode node-specific properties. For example, in the 6-DoF granular system \ref{s_sec:6dof_dem}, scalar labels distinguish original nodes (\( \alpha_i = 0 \)) from ghost boundary-reflection nodes  (\( \alpha_i = 1 \)). In contrast, for the constrained \(N\)-body dataset \ref{s_sec:cons_nbody}, the scalar feature corresponds to each particle’s electric charge (\( \alpha_i = Z_i \)).}

Each edge \({ij}\) is characterized by the edge distance vector between the connected nodes: \(\vec{dx}_{ij} = (\vec{r}_j - \vec{r}_i\)), where \(\vec{r}_j\) and  \(\vec{r}_i\) represent the position vectors of the receiver node $j$ and the sender node $i$ respectively. Additionally, \textbf{edge features} can include scalar labels that encode different types of interactions, such as collision forces, joint constraints, or electromagnetic influences, allowing the model to distinguish and appropriately process the diverse interaction mechanisms present within the multi-body system.

The graph representation, constructed from the system's physical properties, serves as the input data for \textcolor{black}{\textsc{Dynami-CAL GraphNet}}. At each time step \( t \), the model processes the graph and predicts changes in the state of each node, specifically the changes in velocity \(\vec{\delta v}\), angular velocity \(\vec{\delta \omega}\), and position vector \(\vec{\delta r}\). The training data consists of input-output pairs derived from observed trajectories. Each pair comprises the graph representation at time \( t\) and the corresponding changes in state from \( t\) to \( t+1 \). When the positions and angular velocities of the system's components are observed, velocities and angular velocities are computed using finite differences. Alternatively, directly measured linear velocity and angular velocity vectors can be used if available. The vector features of each node, \(\mathbf{V}_i = [\vec{v}_i^t, \vec{\omega}_i^t, \vec{v}_i^{t-1}, \vec{\omega}_i^{t-1}]\), along with the edge distance vector \( \vec{dx}_{ij} \), are normalized by scaling them by their respective maximum magnitudes.
 
\subsection{\textcolor{black}{\textsc{Dynami-CAL GraphNet}} Architecture}
The \textcolor{black}{\textsc{Dynami-CAL}} GraphNet model leverages observed trajectories as training data to learn the system's implicit, edge-wise interaction dynamics. By integrating inductive biases that enforce the conservation of linear and angular momentum, the model ensures the learned dynamics are physically consistent. The model follows \textcolor{black}{the scalarization-vectorization paradigm}, as illustrated in Figure \ref{fig:spatiotemp}.

\textcolor{black}{The scalarization step computes edge embeddings from node and edge features, and the vectorization step decodes these embeddings into }edge-wise forces and moments. \textcolor{black}{These are then aggregated at each node to update its state, and the updated graph then undergoes repeated scalarization–vectorization passes through multiple message-passing iterations, effectively emulating sub-time-step state updates within a single time step. The model predicts node-wise changes in linear velocity (\(\Delta \vec{v}_i\)), angular velocity (\(\Delta \vec{\omega}_i\)), and displacement (\(\Delta \vec{x}_i\)). Training is performed by minimizing the mean squared error between predicted and ground-truth values, with gradients backpropagated to learn the system dynamics.}

\subsubsection{\textcolor{black}{Scalarization}}
In \textcolor{black}{this} step, we transform the vector and scalar properties of nodes and edges into high-dimensional \textcolor{black}{scalar} embeddings. These embeddings serve as a comprehensive representation  of the interactions for each edge, capturing the essential features necessary for the model to understand the system's dynamics.
\subsubsection*{Edge local reference frame calculation}
\label{s_s_sec:refframe}
\begin{figure}[h]  
    \centering  
    \includegraphics[width=0.8\textwidth]{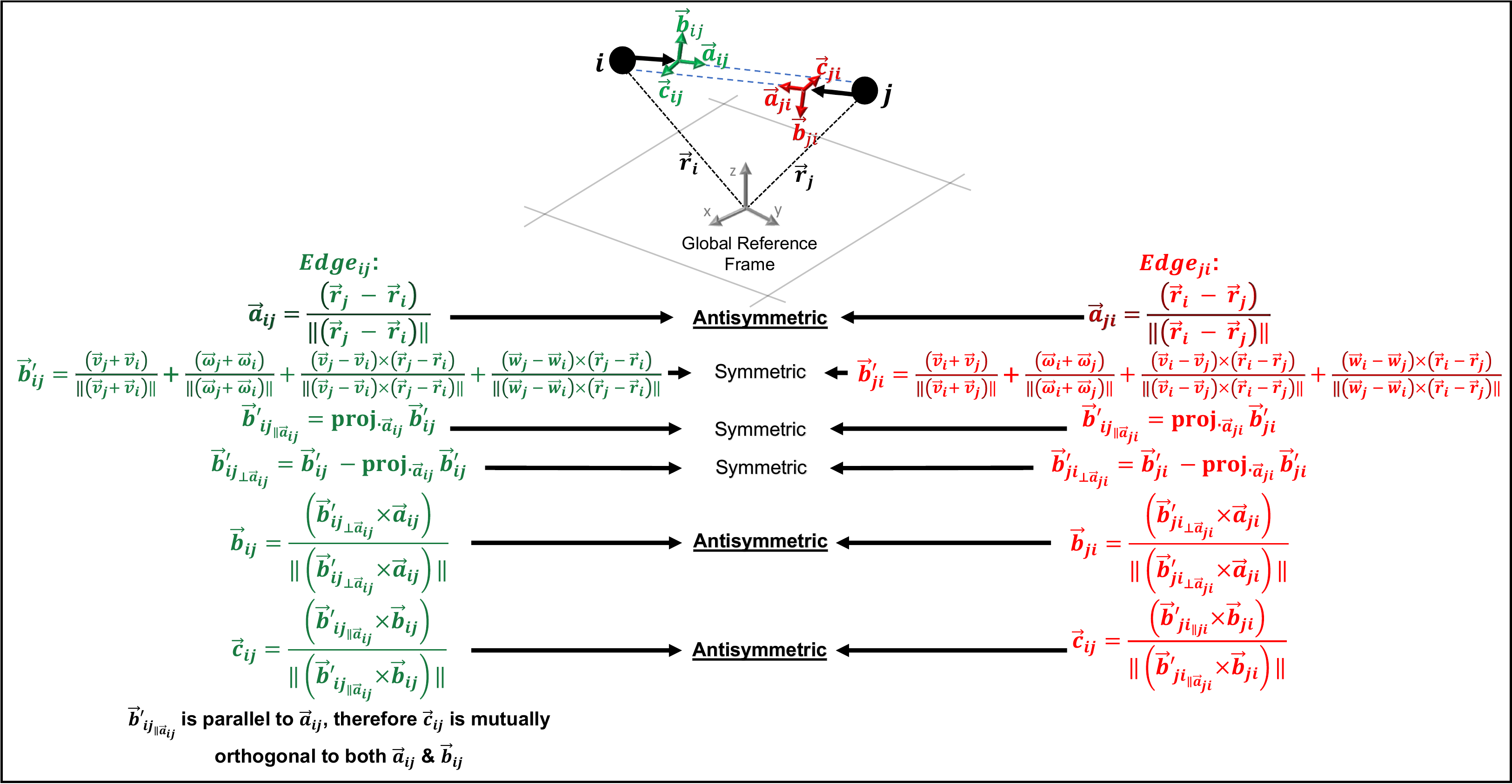}  
    \caption{Permutation equivariant reference frame calculation for bi-directional edges}
    \label{fig:pignn_refframe} 
\end{figure}
Constructing an edge local reference frame--\textcolor{black}{antisymmetric under interchange of nodes--}is crucial for enforcing conservation laws within our model. This process is illustrated in Figure \ref{fig:pignn_refframe}.  For each edge \( e_{ij} \), we begin  by defining the first basis vector \( \vec{a}_{ij} \) as the unit vector along the distance vector between nodes \( i \) and \( j \):
\[
\vec{a}_{ij} = \frac{\vec{r}_j - \vec{r}_i}{\|\vec{r}_j - \vec{r}_i\|}, \text{where } \vec{r}_j \text{ and } \vec{r}_i \text{ are unscaled position vectors of the receiver and sender nodes}
\]
This vector \( \vec{a}_{ij} \) is \textcolor{black}{\textbf{antisymmetric} under node interchange}, meaning that swapping nodes \( i \) and \( j \) reverses its direction. Additionally, \( \vec{a}_{ij} \) is both \textbf{rotation equivariant} and \textbf{translation invariant}.
The challenge lies in constructing the remaining basis vectors, which must form an orthogonal set with \( \vec{a}_{ij} \) while preserving \textcolor{black}{antisymmetry under node interchange}. A straightforward approach, such as computing \( \vec{x}_i \times \vec{x}_j \), initially produces \textcolor{black}{an antisymmetric} vector due to the anti-commutative nature of the cross product. However, deriving a third basis vector through another \textcolor{black}{cross product} results in \textcolor{black}{an unsuitable symmetric} vector.
To address this, we introduce an intermediate vector based on the state vectors of the nodes connected by the edge \(ij\):
\[
\vec{b}_{ij}' = \frac{\vec{v}_j + \vec{v}_i}{\|\vec{v}_j + \vec{v}_i\|} + \frac{\vec{\omega}_j + \vec{\omega}_i}{\|\vec{\omega}_j + \vec{\omega}_i\|} + \frac{(\vec{v}_j - \vec{v}_i) \times (\vec{r}_j - \vec{r}_i)}{\|(\vec{v}_j - \vec{v}_i) \times (\vec{r}_j - \vec{r}_i)\|} + \frac{(\vec{\omega}_j - \vec{\omega}_i) \times (\vec{r}_j - \vec{r}_i)}{\|(\vec{\omega}_j - \vec{\omega}_i) \times (\vec{r}_j - \vec{r}_i)\|}.
\]
This intermediate vector is both \textcolor{black}{\textbf{antisymmetric} to node interchange, }\textbf{translation invariant} and \textbf{rotation equivariant}. We then decompose \( \vec{b}_{ij}' \)  into components parallel and perpendicular to \( \vec{a}_{ij} \):
\[
\vec{b}_{ij\parallel \vec{a}_{ij}}' = \text{proj}_{\vec{a}_{ij}} \vec{b}_{ij}' = \left(\frac{\vec{a}_{ij} \cdot \vec{b}_{ij}'}{\|\vec{a}_{ij}\|^2}\right) \vec{a}_{ij},
\]
\[
\vec{b}_{ij\perp \vec{a}_{ij}}' = \vec{b}_{ij}' - \vec{b}_{ij\parallel \vec{a}_{ij}}'
\]
Both components are \textcolor{black}{antisymmetric to node interchange}.
Using the perpendicular component, we define the second basis vector:
\[
\vec{b}_{ij} = \frac{\vec{b}_{ij\perp \vec{a}_{ij}}' \times \vec{a}_{ij}}{\|\vec{b}_{ij\perp \vec{a}_{ij}}' \times \vec{a}_{ij}\|}
\]
This cross product yields an \textcolor{black}{\textbf{antisymmetric}} vector by combining a \textcolor{black}{symmetric} vector with \textcolor{black}{an antisymmetric} one.
Finally, the third basis vector is computed as:
\[
\vec{c}_{ij} = \frac{\vec{b}_{ij\parallel \vec{a}_{ij}}' \times \vec{b}_{ij}}{\|\vec{b}_{ij\parallel \vec{a}_{ij}}' \times \vec{b}_{ij}\|}
\]
Since \( \vec{b}_{ij\parallel \vec{a}_{ij}}' \) is parallel to \( \vec{a}_{ij} \), the resulting vector \( \vec{c}_{ij} \) is orthogonal to both \( \vec{a}_{ij} \) and \( \vec{b}_{ij} \), while also maintaining \textcolor{black}{antisymmetry under node interchange}.

By constructing this orthogonal basis set  \( \vec{a}_{ij}, \vec{b}_{ij}, \vec{c}_{ij} \)-- \textcolor{black}{\textbf{antisymmetric under node interchange, translation invariant,} and rotation equivariant}-- the model ensures symmetrical interactions, which are vital for enforcing the conservation of linear and angular momentum.
\paragraph{Degeneracy of the Local Reference Frame}  
The local reference frame becomes degenerate under two specific conditions: 1) When the intermediate vector \(\vec{b}_{ij}'=0 \): In this scenario, the edge system is stationary, exhibiting no linear or angular velocities. The interaction can be fully captured using only the first basis vector \(\vec{a}_{ij}\) along the edge. 2) When \(\vec{b}_{ij}'\) is parallel to \(\vec{a}_{ij}\): This indicates that the velocities and angular velocities are aligned with the edge vector, implying that the interaction is constrained along the edge direction. Consequently, \(\vec{a}_{ij}\) sufficiently represents the interaction. In both cases, the system remains \textbf{effectively non-degenerate} for representing the relevant interactions, ensuring robust and accurate modeling of the multi-body system's dynamics.

\subsubsection*{\textcolor{black}{Scalar Edge Embedding from Projections onto Edge-Reference Frames}}
\label{s_s_sec:interaction_embedding_encoding}
After establishing the edge-wise local reference frames, the vector features of the connected nodes are projected onto these reference frames. Specifically, for an edge \( ij \), the  sender node's vector features are defined as:
\(
\mathbf{V}_i = [\vec{v}_i^t, \vec{\omega}_i^t, \vec{v}_i^{t-1}, \vec{\omega}_i^{t-1}]
\). These features are projected onto the basis vectors of the edge's reference frame \( \vec{a}_{ij}, \vec{b}_{ij}, \vec{c}_{ij} \). Conversely, for the receiver node \( j \), its vector features \( \mathbf{V}_j \) are projected onto the \textcolor{black}{\textbf{antisymmetric}} reference frame, specifically \( -\vec{a}_{ij}, -\vec{b}_{ij}, -\vec{c}_{ij} \).
This projection strategy ensures that the scalar projections for the sender (\( i \)) and receiver (\( j \)) nodes remain invariant when  the nodes are swapped. To illustrate, consider the reverse direction edge \( ji \):
\begin{itemize}
    \item The sender node \( j \)'s features \(\vec{a}_{ji}, \vec{b}_{ji}, \vec{c}_{ji} \) are projected onto the reference frame of edge \( ji \) , which corresponds  to \( -\vec{a}_{ij}, -\vec{b}_{ij}, -\vec{c}_{ij} \).
    \item The receiver node \( i \)'s features are then projected onto the \textcolor{black}{antisymmetric} reference frame \(-\vec{a}_{ji}\), \(-\vec{b}_{ji}\), \(-\vec{c}_{ji} \), which is equivalent to \( \vec{a}_{ij} \), \(\vec{b}_{ij} \), \(\vec{c}_{ij} \).
\end{itemize}
This ensures that node \( i \)'s vectors are always aligned with the reference frame of edge \( ij \), and node \( j \)'s vectors are aligned with edge \( ji \), regardless of the edge direction. By maintaining this structure, we \textcolor{black}{achieve \textbf{node-interchange invariant}} scalar features for constructing interaction embedding \textcolor{black}{for edges that remain invariant to node interchange} \(^{1}\). 

Furthermore, the projected scalars -- and thus the resulting interaction embeddings -- inherit additional symmetries based on the properties of the edge reference frame. Specifically, if the reference frame is rotation equivariant, the projected scalars remain \textbf{rotation invariant}. This is because the relative alignment between vectors and the basis vectors stays consistent under rotation. Similarly, since the state vectors (e.g., velocity and angular velocity) are \textbf{translation invariant}, the projected scalars inherit translation invariance provided the reference frame is translation invariant.

These projected scalars for both sender and receiver nodes are transformed into higher-dimensional embeddings, denoted as \( \epsilon_{ij}^{\text{sender}} \) and \( \epsilon_{ij}^{\text{receiver}} \), using the function \( \phi_{e_1} \), which is implemented as a multi-layer perceptron (MLP): \(
\epsilon_{ij}^{\text{sender}} = \phi_{e_1}\left(\text{proj}_{\text{frame}} \mathbf{V}_i\right), \quad \epsilon_{ij}^{\text{receiver}} = \phi_{e_1}\left(\text{proj}_{\text{frame}} \mathbf{V}_j\right)
\).

Additionally, we create \textcolor{black}{another invariant} embedding from the \textcolor{black}{magnitude of the} edge distance vector \( \Delta \vec{x}_{ij} = \vec{r}_j - \vec{r}_i \) using another  MLP \( \phi_{e_2} \):
\(
\epsilon_{ij}^{edge} = \phi_{e_2}(||\Delta \vec{x}_{ij}||)
\).

The node scalar features \( \alpha_i \)  for each node \( v_i \in G(V, E) \) are encoded using the MLP function \( \phi_n \): \(
h_i = \phi_n(\alpha_i)\). For an edge \( ij \), the node embeddings \( h_i \) and \( h_j \) correspond to nodes \( i \) and \( j \), respectively.

The edge embeddings -- \(\epsilon_{ij}^{edge}\), \(\epsilon_{ij}^{sender}\), \(\epsilon_{ij}^{receiver}\) -- along with node embeddings \( h_j \), \( h_j \), are then processed in the subsequent step to create \textcolor{black}{the final comprehensive and expressive} edge interaction embedding.
\footnotetext[1]{\textbf{Intuition for \textcolor{black}{Invariance of Edge Embedding under Node Interchange}:}
\textcolor{black}{Invariance of edge embedding under node-interchange} is crucial for physically consistent modeling, especially in systems where interactions depend on the relative positions or states of nodes, such as forces in spring or other pairwise interactions. For instance, consider a spring connecting two nodes \( i \) and \( j \) with positions \( \vec{r}_i \) and \( \vec{r}_j \). The \textbf{stretch} or \textbf{compression} of the spring depends solely on the relative distance \( \|\vec{r}_j - \vec{r}_i\| \), which remains unchanged regardless of the order in which the nodes are considered. Therefore, to accurately model such physical interactions, the embeddings derived from the node features must be invariant. In our context, this means that the interaction embeddings for edges \( ij \) and \( ji \) are identical, ensuring consistency and physical accuracy in the model's predictions.}
\subsubsection*{Final Edge Interaction Embedding}
\label{s_s_sec:interaction_embedding_procesing}
In this step, the edge embeddings are first combined and then transformed using a function \( \theta_e \) to produce an invariant edge embedding:
\[
\epsilon_{ij} = \theta_e\left(\epsilon_{ij}^{\text{edge}} + \epsilon_{ij}^{\text{sender}} + \epsilon_{ij}^{\text{receiver}} + h_i + h_j\right)
\]
\subsubsection*{\textcolor{black}{Incorporating Interaction History in Edge Embedding via Skip Connection}}
The interaction embedding at the message-passing step \(L\), denoted \( \epsilon_{ij}^{L} \), is combined with the previous layer's edge embedding \( \epsilon_{ij}^{L-1} \) through a skip connection. The resulting sum is passed through a layer normalization operation (\(\theta_{\text{LN}}\)) to yield the updated interaction embedding \( \epsilon_{ij}' \). This recursive dependence on prior edge embeddings enables the model to capture temporal dynamics unfolding across multiple steps. At each message-passing step, the raw interaction embedding \( \epsilon_{ij}^{L} \) is enriched with spatial context through distance-based features derived from neighboring node configurations, while the skip connection integrates temporal memory from previous interactions. Together, these mechanisms allow the model to learn localized physical interactions across both space and time, leading to robust and generalizable predictions in a wide range of structured dynamical systems.

\subsubsection{\textcolor{black}{Vectorization}}
In the \textcolor{black}{vectorization} step, the invariant edge interaction embedding \( \epsilon_{ij}' \) is decoded using multiple MLP functions to extract the internal forces \( \vec{F}_{ij} \) and rotational torques \( \vec{\tau}_{ij} \) vectors, while ensuring the conservation of both linear and angular momentum. These vectors are then aggregated for each node to account for the cumulative effects of all interactions.

Additionally, this step involves estimating the inverse mass \( \frac{1}{m_i} \) and inverse moment of inertia \( \frac{1}{I_i} \) for each node from their scalar embeddings. These values are subsequently used to compute the change in linear velocity \( \Delta \vec{v}_i \) and angular velocity \( \Delta \vec{\omega}_i \), enabling accurate updates to the system's dynamics.

If external forces are present, the changes in velocity and angular velocity are decoded directly from the node scalar embeddings \( h_i \) for each node \( v_i \). This allows the model to incorporate both internal interactions and external influences in a physically consistent manner.
\subsubsection*{Internal Force Vectors}
The invariant edge interaction embedding \( \epsilon_{ij}' \) is decoded into invariant coefficients which modulate the reference frame basis vectors \( \vec{a}_{ij}, \vec{b}_{ij}, \vec{c}_{ij} \). This results in the internal forces \( \vec{F}_{ij} \) as follows:
\[
\vec{F}_{ij} = \psi_{e_f}(\epsilon_{ij}')[0] \cdot \vec{a}_{ij} + \psi_{e_f}(\epsilon_{ij}')[1] \cdot \vec{b}_{ij} + \psi_{e_f}(\epsilon_{ij}')[2] \cdot \vec{c}_{ij}
\]
Here, \( \psi_{e_f}(\epsilon_{ij}') \) provides the scalar coefficients for the basis vectors. By construction, the forces are anti-symmetric, ensuring the conservation of linear momentum.
\[
\vec{F}_{ij} = -\vec{F}_{ji}
\]
This anti-symmetry guarantees that the internal forces between any two connected nodes $i$ and $j$ are equal in magnitude and opposite in direction, maintaining the physical principle of \textbf{linear momentum conservation} within the system.

\subsubsection*{Rotational Torque Vectors: Isolated Edge Dynamical System}
\textcolor{black}{Rotational torque is decoded by enforcing angular momentum conservation at each edge, modeled as an isolated dynamical system. We begin by demonstrating conservation about a reference point \( \vec{r}_0 \) for two interacting bodies \( i \) and \( j \), each with velocity \( \vec{v}_i \), angular velocity \( \vec{\omega}_i \), mass \( m_i \), and moment of inertia \( I_i \).}

\textcolor{black}{\paragraph{Conservation of Angular Momentum for Two Interacting Bodies}}

\textcolor{black}{The angular momentum of body \( i \) about a reference point \( \vec{r}_0 \) comprises both spin and orbital components:
\begin{equation}
L_i^t = I_i \vec{\omega}_i^t + (\vec{r}_i^{\,t} - \vec{r}_0) \times m_i \vec{v}_i^{\,t}.
\label{eq:L_single_body}
\end{equation}}

\textcolor{black}{In a closed two-body system, total angular momentum is then given by:
\begin{equation}
L^t = L_i^t + L_j^t
\end{equation}
Conservation of angular momentum implies: 
\begin{equation}
L^{t+\delta t} = L^t,
\label{eq:ang_mom_conservation}
\end{equation}
which in turn implies that the total angular momentum transfer from body \( i \) to \( j \), denoted \( \vec{A}_{ij} \), is equal and opposite to that from \( j \) to \( i \), denoted \( \vec{A}_{ji} \).
The expression for \( \vec{A}_{ij} \) is:
\begin{equation}
\vec{A}_{ij} = I_i \left(\vec{\omega}_i^{t+\Delta t} - \vec{\omega}_i^{\,t}\right) + (\vec{r}_i^{\,t} - \vec{r}_0) \times \vec{F}_{ij},
\label{eq:Aij}
\end{equation}}

\textcolor{black}{The rotational torque that contributes to updating the spin angular velocity of body \( i \) is computed by decoupling the spin component from \( \vec{A}_{ij} \). The resulting expression for the change in spin angular momentum of body \( i \) is:
\begin{equation}
I_i \left(\vec{\omega}_i^{\text{final}} - \vec{\omega}_i^{\text{initial}}\right) = \vec{A}_{ij} - (\vec{r}_i - \vec{r}_{0_{ij}}) \times \vec{F}_{ij},
\label{eq:rotational_momentum_i}
\end{equation}}

\textcolor{black}{and a similar expression holds for the rotational torque acting on body \( j \).
For the full derivation, refer to Supplementary Information Section~\S5.4.}

\textbf{Conservation of Angular Momentum in \textsc{Dynami-CAL} GraphNet:} For any edge \( ij \) connecting nodes \( i \) and \( j \), the internal angular interaction vector \( \vec{A}_{ij} \) is decoded from the edge interaction embedding \( \epsilon_{ij}' \) using the function \( \psi_{e_a} \). The invariant edge interaction embedding \( \epsilon_{ij}' \) is transformed into scalar coefficients, which are then \textcolor{black}{used to modulate  the basis vectors of the local }reference frame \( \vec{a}_{ij}, \vec{b}_{ij}, \vec{c}_{ij} \):
\[
\vec{A}_{ij} = \psi_{e_a}(\epsilon_{ij}')[0] \cdot \vec{a}_{ij} + \psi_{e_a}(\epsilon_{ij}')[1] \cdot \vec{b}_{ij} + \psi_{e_a}(\epsilon_{ij}')[2] \cdot \vec{c}_{ij}
\]
\textcolor{black}{Since the basis vectors are antisymmetric under node interchange, the decoded interaction vector \(\vec{A}_{ij}\) also preserves this antisymmetry:
\begin{equation}
\vec{A}_{ij} = -\vec{A}_{ji},
\end{equation}
thereby ensuring angular momentum conservation as stated in Equation~\ref{eq:ang_mom_conservation}.}

\textcolor{black}{To isolate the spin component from \(\vec{A}_{ij}\), we compute the reference point—also referred to as the \textbf{point of action}—for the internal force. This point, denoted \( \vec{r}_{0_{ij}} \), is computed as a weighted sum of the position vectors of nodes \( i \) and \( j \).
The weights are derived from the node scalar embeddings \( h_i \) and \( h_j \) using the function \( \psi_{n1} \).}
\[
\vec{r}_{0_{ij}} = \frac{\psi_{n1}(h_i) \cdot \vec{r}_i + \psi_{n1}(h_j) \cdot \vec{r}_j}{\psi_{n1}(h_i) + \psi_{n1}(h_j)}
\]
This formulation ensures that the reference point remains  consistent \textcolor{black}{under node interchange for bi-directional edges}:
\[
\vec{r}_{0_{ij}} = \vec{r}_{0_{ji}}.
\]
After \( \vec{F}_{ij} \), \( \vec{A}_{ij} \), and \( \vec{r}_{0_{ij}} \) are decoded for edge \( ij \), the \textbf{rotational torque} on the receiver node \( j \) is computed as:
\[
I_j \cdot \Delta \vec{\omega}_j = \vec{A}_{ij} - (\vec{r}_{0_{ij}} - \vec{r}_j) \times \vec{F}_{ij}\cdot \lambda_{ij}.
\]
Here \(\lambda_{ij}=\psi_{e_l}(\epsilon_{ij}')\) represents a scalar decoded from the edge interaction embedding, introduced to enhance stability by mitigating the influence of negligible noisy edge forces on the calculation of rotational torque. This approach ensures that the predicted rotational torques between nodes are \textbf{physically consistent} (Physics derivation shown in Equation \ref{eq:rotational_momentum_i}), thereby upholding the conservation of angular momentum throughout the system.
\subsubsection*{Aggregation of edge forces and moments on the nodes}
The decoded forces and moments from each edge are aggregated on the receiver nodes. These aggregated internal forces and moments are then used to determine the changes in linear and angular velocities for each node.
\subsubsection*{Decoding Change in Velocity and Angular Velocity for Each Node}
Using the functions \( \psi_{n2} \) and \( \psi_{n3} \), the inverse mass \( \frac{1}{m_i} \) and inverse moment of inertia \( \frac{1}{I_i} \) are decoded from each node's scalar embeddings \( h_i \). These decoded values are utilized to compute the changes in linear velocity \( \Delta \vec{v}_i \) and angular velocity \( \Delta \vec{\omega}_i \) for each node:
\[
\Delta \vec{v}_i = \psi_{n2}(h_i) \cdot \sum \vec{F}_{ij}, \quad \Delta \vec{\omega}_i = \psi_{n3}(h_i) \cdot \sum \vec{M}_{ij}
\]
where \( \sum \vec{F}_{ij} \) represents the total internal force acting on node \( i \) from all connected edges and \( \sum \vec{M}_{ij} \) represents the total internal torque acting on node \( i \).
These updates ensure accurate changes in the system's dynamics based on both internal interactions and node-specific properties.
\subsubsection*{Decoding external forces}
\label{s_s_sec:ext_force_decoder}
Additionally, when external forces are present, the change in velocity due to external influences is decoded using the function \( \psi_{n4} \):
\[
\Delta \vec{v}_i^{\text{ext}} = \psi_{n4}(h_i)
\]
\subsubsection*{\textcolor{black}{Updating the Graph State}}
Finally, the net change in both linear and angular velocities, resulting from internal and external forces, is computed and applied to update the node states. This step is crucial for advancing the system dynamics forward in time.
The net change in linear velocity is computed as:
\[
\Delta \vec{v}_i^{\text{net}} = \Delta \vec{v}_i + \Delta \vec{v}_i^{\text{ext}}
\]
Thus, the updated velocity of node \( i \) is:
\[
\vec{v}_i^{\text{new}} = \vec{v}_i + \Delta \vec{v}_i^{\text{net}}
\]
Similarly, the net change in angular velocity is given by:
\[
\Delta \vec{\omega}_i^{\text{net}} = \Delta \vec{\omega}_i
\]
resulting in the updated angular velocity:
\[
\vec{\omega}_i^{\text{new}} = \vec{\omega}_i + \Delta \vec{\omega}_i^{\text{net}}
\]
Subsequently, the position of each node is updated using the computed velocities through single-step Euler integration, utilizing  the same time step \( \Delta t \) as used  during forward differencing:
\[
\Delta \vec{x}_i = \frac{(\vec{v}_i + \vec{v}_i^{\text{new}})}{2} \Delta t
\]
Thus, the new position of node \( i \) is:
\[
\vec{x}_i^{\text{new}} = \vec{x}_i + \Delta \vec{x}_i
\]
The updated system state is then fed back into the model pipeline, starting from the encode step, allowing for iterative updates. This iterative process ensures that both velocities and positions are adjusted based on the cumulative effects of internal and external forces. By incorporating forward integration bias, the method achieves physically consistent multi-step updates, enabling precise and interpretable modeling of the evolving system dynamics over time.
\begin{figure}[h]  
    \centering  
    \includegraphics[width=0.9\textwidth]{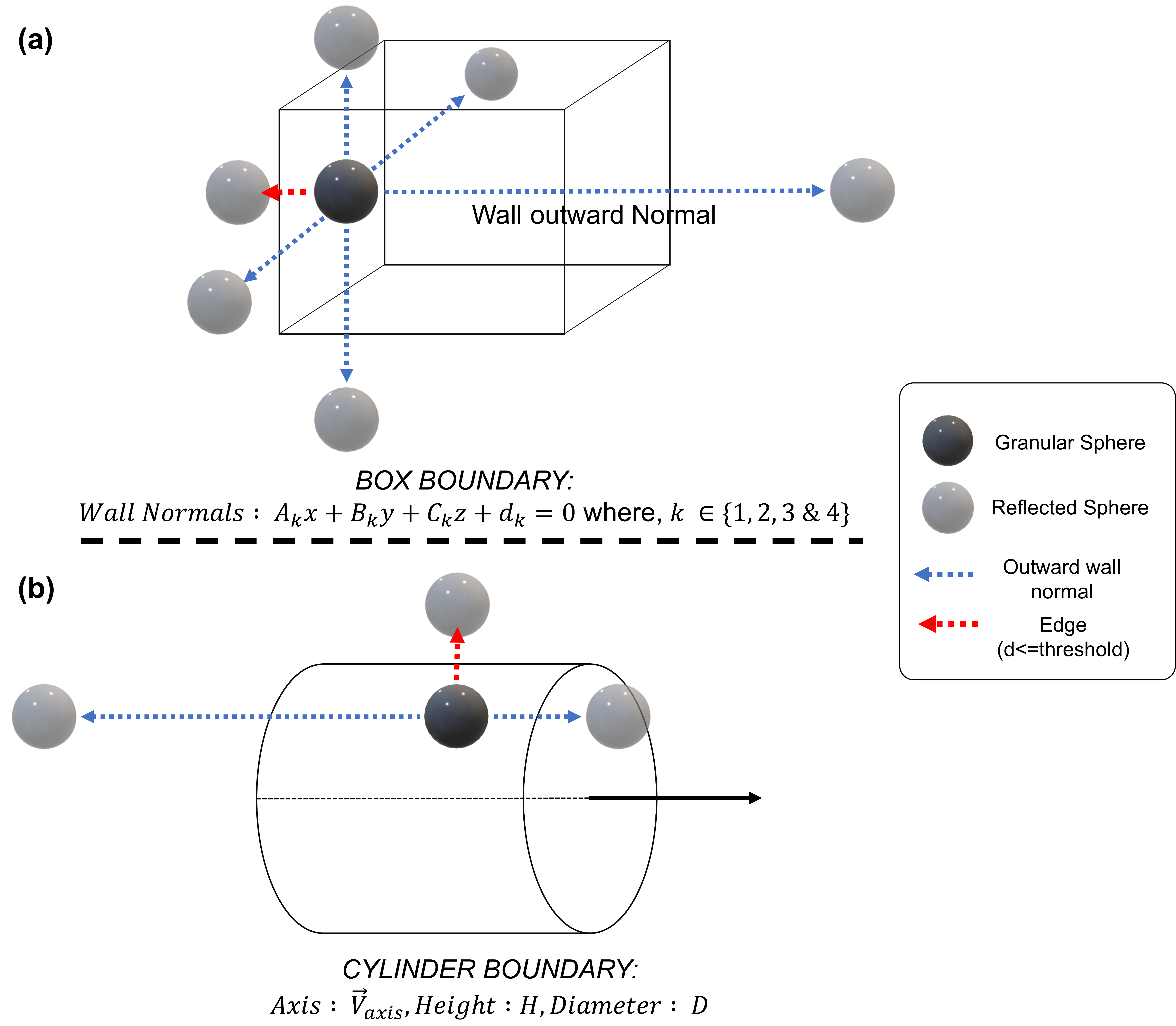}  
    \caption{\textbf{Mesh and particle-free modeling of wall boundaries}. (a) Illustrates the normal contact and collision modeling of a granular sphere with the walls of a box boundary. The sphere is reflected along the outward normal vector of each wall. Of the six possible reflections, only those that satisfy a predefined threshold distance form an edge with the reflected sphere.(b) Demonstrates the contact modeling for a cylindrical boundary, which is parameterized by its diameter, height, and axis vector. The spheres are reflected off both the curved surface and the planar end caps, effectively handling interactions with the cylindrical geometry.}
    \label{fig:pignn_boundary}  
\end{figure}
\subsubsection{Mesh-particle free modeling of boundaries through reflections}
\label{s_s_sec:wall_model}
Many dynamical systems involve interactions between their components and boundaries. Such systems are prevalent in various domains, including granular systems (as demonstrated in Section \ref{sec:results}), rigid body dynamics (e.g., spheres rolling on a surface), and musculoskeletal systems. \textcolor{black}{In this work, we introduce a mesh-free and particle-free approach for modeling boundary interactions in multi-body dynamical systems. Unlike prior methods that rely on explicit meshes or densely sampled particles to represent walls and enclosures \cite{satorras2021n, pfaff2020learning, allen2022learning, allen2023graph}, \textsc{Dynami-CAL GraphNet} models each boundary as a reflective surface defined by its outward normal. Physical components are mirrored across this normal to generate ghost nodes that encode boundary effects (Figure~\ref{fig:pignn_boundary}). Crucially, these boundary interactions are integrated into the message-passing framework in the same manner as body–body interactions, leveraging consistent geometric structures without requiring task-specific modules or special heuristics.}

The reflection process leverages the outward normal to the boundary. For a body at position $\vec{r}$, the reflected position $\vec{r}_{\text{reflected}}$ is computed using the outward normal vector $\vec{n}$ as follows:
\[
\vec{r}_{\text{reflected}} = \vec{r} - 2 (\vec{r} \cdot \vec{n}) \vec{n}.
\]
where $\vec{n}$ is the outward normal vector to the boundary. \textcolor{black}{For planar boundaries such as floors or walls, this results in one reflected node per physical node per wall. For example, a single floor yields \( N_n \) ghost nodes, while a cuboidal enclosure with six walls produces \( 6N_n \) ghost nodes. In the case of a curved cylindrical boundary, the normal \( \vec{n} \) is computed by normalizing the vector from the cylinder’s axis to the particle’s position, ensuring radially outward reflection. If the cylinder is capped, two additional planar reflections are applied, resulting in a total of \( 3N_n \) ghost nodes. As previously discussed, this introduces at most \( (W{-}1) \times N_n \) additional nodes for \( W \) boundaries, a fixed and tractable overhead that remains significantly lower than particle- or mesh-based representations \cite{allen2022learning}, and does not scale with boundary size.}

The reflected body inherits wall-specific features, including velocity and angular velocity vectors, as well as one-hot encoded labels indicating blocked degrees of freedom. For stationary walls, both velocity and angular velocity vectors are set to zero, ensuring an accurate physical representation of boundary constraints. \textcolor{black}{For moving boundaries, the reflected (ghost) nodes inherit motion characteristics from the wall, including translational velocity, tangential velocity induced by rotation, and angular velocity—depending on the wall’s dynamic state. These quantities are computed directly from the wall’s geometry and motion. For instance, in the case of a rotating cylindrical boundary (Figure~\ref{fig:pignn_boundary}), the tangential velocity of each reflected node is calculated using the wall’s angular velocity vector \( \vec{\omega} \) and the reflected node’s position relative to the cylinder’s axis of rotation, denoted \( \vec{d}_{\text{reflected}, i} \). The tangential velocity is given by:
\[
\vec{v}_{\text{tangential}} = \vec{d}_{\text{reflected}, i} \times \vec{\omega}.
\]}

\textcolor{black}{This formulation ensures that ghost nodes inherit the correct dynamic boundary conditions, allowing the model to capture the influence of both translational and rotational wall motion in a physically consistent and unified manner.}

\textcolor{black}{The system of physical spheres and their reflections is represented as a graph, where nodes correspond to both real bodies and their ghost counterparts. Edges are dynamically constructed at each time step between a physical node and its reflected ghost node if their separation distance falls below a threshold proportional to the particle’s diameter. This formulation ensures that only bodies in close proximity to a boundary form interactions with their reflections, enabling accurate modeling of boundary contact effects while keeping computational overhead minimal.}

\textcolor{black}{During rollout,} this reflection process is recalculated at every time step, ensuring that interactions remain strictly aligned along the normal direction. 

\section*{Ethics Statement}

This paper presents \textsc{Dynami-CAL GraphNet}, a physics-informed, learning-based method for modeling discrete dynamical systems.  
All experiments use either synthetic simulations or publicly available benchmarks that do not include any personally identifiable or sensitive information; thus, no human-subject approval was required.  
We advocate for the responsible use of this technology, especially in real-world or safety-critical applications, where rigorous validation and domain-specific safeguards are essential.

\section*{Data Availability}
The dataset for the 6-DoF granular collision benchmark was generated using the MFIX-DEM simulator, available at \url{https://mfix.netl.doe.gov/products/mfix/}.  The exact simulation parameters are detailed in Supplementary Information Section~1. The dataset will be made publicly available upon acceptance.
The constrained N-body and human motion benchmarks were obtained from the GMN~\cite{huang2022equivariant} repository: \url{https://github.com/hanjq17/GMN}.  
The protein dynamics benchmark was sourced from the EGHN~\cite{han2022equivariant} repository: \url{https://github.com/hanjq17/EGHN}.

\section*{Code Availability}
The proposed method is implemented in PyTorch. The code is currently under patent review and will be made publicly available upon acceptance.

\section*{Acknowledgments}

 This research was funded by the Swiss National Science Foundation (SNSF) Grant Number 200021\_200461.

\bibliographystyle{ieeetr} 
\bibliography{references}

\end{document}